\pdfoutput=1
\documentclass[11pt]{article}

\usepackage[]{naacl2021}

\usepackage{times,latexsym}
\usepackage[T1]{fontenc}
\usepackage{makecell} % not sure this is needed
\usepackage{multirow} 
\usepackage{multicol}
\usepackage{makecell}
\usepackage{tabularx}
\usepackage{color}
\usepackage{xcolor}
\usepackage{listings}
\usepackage{subcaption}
\usepackage{url}
\usepackage{latexsym}
\usepackage[pdftex]{graphicx}
\usepackage[inline]{enumitem}
\usepackage{array}
\usepackage{setspace}
\usepackage{framed}
\usepackage{xspace}
\usepackage{amsmath}
\usepackage{amssymb}
\usepackage{cleveref}
\usepackage{booktabs}
\usepackage{colortbl}
\usepackage{lscape}
\usepackage{adjustbox}
\usepackage{rotating}
\usepackage{arydshln}
\usepackage{soul}
\usepackage[normalem]{ulem} 
\usepackage{subcaption}
\usepackage{soul}
\usepackage{tikz}
\usetikzlibrary{calc}
\usepackage{amsmath}
\usepackage{xcolor}
\usepackage{diagbox}
\usepackage{todonotes}

\usepackage{arydshln}
\usepackage{afterpage}

\newcounter{srCounter}
\newif\ifsrvar
\srvartrue
\ifsrvar
\newcommand{\seb}[1]{{\small \color{red} \refstepcounter{srCounter}\textsf{[SR]$_{\arabic{srCounter}}$:{#1}}}}
\else
\newcommand{\seb}[1]{}
\fi

\newcounter{fpCounter}
\newif\iffpvar
\fpvartrue
\iffpvar
\newcommand{\fabio}[1]{{\small \color{blue} \refstepcounter{fpCounter}\textsf{[FP]$_{\arabic{fpCounter}}$:{#1}}}}
\else
\newcommand{\fabio}[1]{}
\fi

\newcounter{trCounter}
\newif\iftrvar
\trvartrue
\iftrvar
\newcommand{\tim}[1]{{\small \color{purple} \refstepcounter{trCounter}\textsf{[TR]$_{\arabic{trCounter}}$:{#1}}}}
\else
\newcommand{\tim}[1]{}
\fi

\newcounter{apCounter}
\newif\ifapvar
\apvartrue
\ifapvar
\newcommand{\piktus}[1]{{\small \color{orange} \refstepcounter{apCounter}\textsf{[AP]$_{\arabic{apCounter}}$:{#1}}}}
\else
\newcommand{\piktus}[1]{}
\fi

\newcounter{plCounter}
\newif\ifplvar
\plvartrue
\ifplvar
\newcommand{\patrick}[1]{{\small \color{magenta} \refstepcounter{plCounter}\textsf{[PL]$_{\arabic{plCounter}}$:{#1}}}}
\else
\newcommand{\patrick}[1]{}
\fi

\newcounter{afCounter}
\newif\ifafvar
\afvartrue
\ifafvar
\newcommand{\angela}[1]{{\small \color{olive} \refstepcounter{afCounter}\textsf{[AF]$_{\arabic{afCounter}}$:{#1}}}}
\else
\newcommand{\angela}[1]{}
\fi

\newcounter{jtCounter}
\newif\ifjtvar
\jtvartrue
\ifjtvar
\newcommand{\jt}[1]{{\small \color{purple} \refstepcounter{jtCounter}\textsf{[JT]$_{\arabic{jtCounter}}$:{#1}}}}
\else
\newcommand{\jt}[1]{}
\fi

\makeatletter
\def\adl@drawiv#1#2#3{%
        \hskip.5\tabcolsep
        \xleaders#3{#2.5\@tempdimb #1{1}#2.5\@tempdimb}%
                #2\z@ plus1fil minus1fil\relax
        \hskip.5\tabcolsep}
\newcommand{\cdashlinelr}[1]{%
  \noalign{\vskip\aboverulesep
           \global\let\@dashdrawstore\adl@draw
           \global\let\adl@draw\adl@drawiv}
  \cdashline{#1}
  \noalign{\global\let\adl@draw\@dashdrawstore
           \vskip\belowrulesep}}
\makeatother

\colorlet{punct}{red!60!black}
\definecolor{background}{HTML}{EEEEEE}
\definecolor{delim}{RGB}{20,105,176}
\colorlet{numb}{magenta!60!black}

\lstdefinelanguage{json}{
    basicstyle=\normalfont\ttfamily,
    numbers=left,
    numberstyle=\scriptsize,
    stepnumber=1,
    numbersep=8pt,
    showstringspaces=false,
    breaklines=true,
    frame=lines,
    backgroundcolor=\color{background},
    literate=
     *{0}{{{\color{numb}0}}}{1}
      {1}{{{\color{numb}1}}}{1}
      {2}{{{\color{numb}2}}}{1}
      {3}{{{\color{numb}3}}}{1}
      {4}{{{\color{numb}4}}}{1}
      {5}{{{\color{numb}5}}}{1}
      {6}{{{\color{numb}6}}}{1}
      {7}{{{\color{numb}7}}}{1}
      {8}{{{\color{numb}8}}}{1}
      {9}{{{\color{numb}9}}}{1}
      {:}{{{\color{punct}{:}}}}{1}
      {,}{{{\color{punct}{,}}}}{1}
      {\{}{{{\color{delim}{\{}}}}{1}
      {\}}{{{\color{delim}{\}}}}}{1}
      {[}{{{\color{delim}{[}}}}{1}
      {]}{{{\color{delim}{]}}}}{1},
}

\usepackage{microtype}

\usepackage{xspace,mfirstuc,tabulary}

\title{KILT: a Benchmark for Knowledge Intensive Language Tasks}

\newcommand{\fair}{$^1$}
\newcommand{\loria}{$^3$}
\newcommand{\ucl}{$^2$}
\newcommand{\uclfair}{$^{1,2}$}
\newcommand{\loriafair}{$^{1,3}$}
\newcommand{\cambridge}{$^4$}
\newcommand{\hf}{$^5$}
\newcommand{\uva}{$^6$}

\author{
Fabio Petroni\fair{} \
Aleksandra Piktus\fair{} \
Angela Fan\loriafair{} \
Patrick Lewis\uclfair{} \\
{\bf Majid Yazdani\fair{} \ Nicola De Cao\uva \  James Thorne\cambridge{} \ Yacine Jernite\hf{} \ Vladimir Karpukhin\fair{} } \\ 
{\bf Jean Maillard\fair{} \ Vassilis Plachouras\fair{} \ Tim Rockt\"aschel\uclfair{} \ Sebastian Riedel\uclfair{} } \\
\fair{}Facebook AI Research \ \ucl{}University College London \ \loria{}LORIA\\
\cambridge{}University of Cambridge \ \hf{}HuggingFace \ \uva{}University of Amsterdam
}

\date{}

\begin{document}
\maketitle
\begin{abstract}
 Challenging problems such as open-domain question answering, fact checking, slot filling and entity linking require access to large, external knowledge sources.  
While some models do well on individual tasks, developing general models 
is difficult as each task might require computationally expensive indexing of custom knowledge sources, in addition to dedicated infrastructure.
To catalyze research on models that condition on specific information in large textual resources, we present a benchmark for knowledge-intensive language tasks~(KILT). 
All tasks in KILT are grounded in the same snapshot of Wikipedia, reducing engineering turnaround through the re-use of components, as well as accelerating research into task-agnostic memory architectures.
We test both task-specific and general baselines, 
evaluating downstream performance in addition to the ability of the models to provide provenance.
We find that a shared dense vector index coupled with a seq2seq model is a strong baseline, outperforming more tailor-made approaches for fact checking, open-domain question answering and dialogue, and yielding competitive results on entity linking and slot filling, by generating disambiguated text. KILT data and code are available at \url{https://github.com/facebookresearch/KILT}.\footnote{and at \url{https://huggingface.co/datasets?search=kilt}}
\end{abstract}

\section{Introduction}

%\fabio{NOTE: when we talk about KILT knowledge, it's "KILT knowledge source", but if we are speaking about wikipedia in general, we do "wikipedia snapshot"}

%\fabio{NOTE3: please don't remove content from the paper when cutting it short, just comment it out}

\begin{figure*}[t!]
    \centering
    \includegraphics[width=\linewidth]{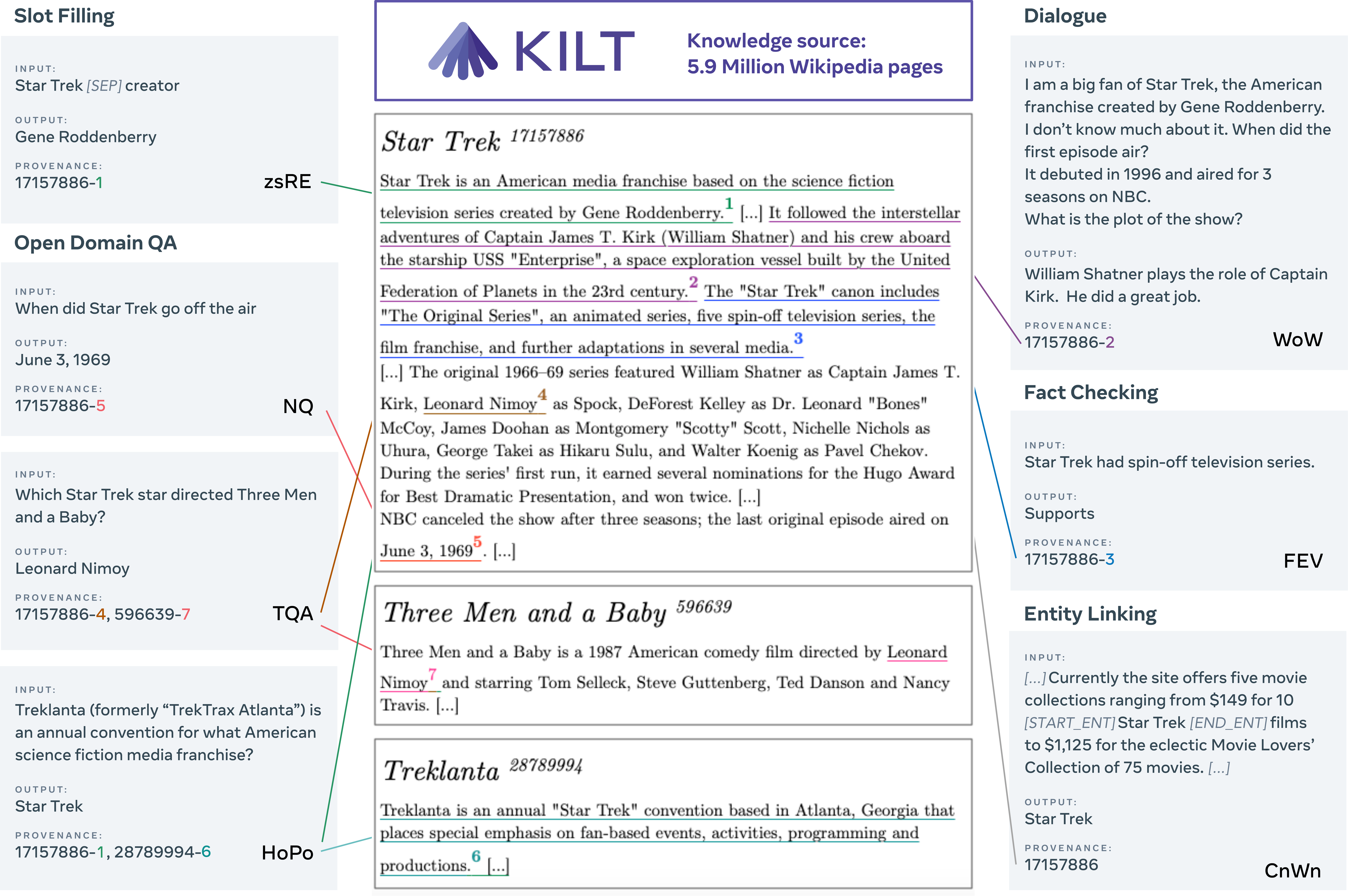}
    \caption{Common KILT interface for knowledge intensive language tasks: each instance consists of \textit{input} and \textit{output} with a \textit{provenance} (text span) from the common KILT knowledge source. %. The latter comes in the form of a textual span within an article in the KILT knowledge source. 
    Source: \texttt{https://en.wikipedia.org/wiki/\{Star\_Trek,Three\_Men\_and\_a\_Baby,Treklanta\}
    }}
    
    \label{fig:example}
\end{figure*}

There has been substantial progress on natural language processing tasks where the inputs are short textual contexts such as a sentences, paragraphs, or perhaps a handful of documents. Critically, we have seen the emergence of general-purpose architectures and pre-trained models that can be applied to a wide range of such tasks~\cite{devlin-etal-2019-bert}.  However, for many real world problems, processing at this local level is insufficient. For example, in open-domain question answering~\cite{chen2017reading} models need to find answers within a large corpus of text. Fact checking a claim~\cite{Thorne18Fever} requires models to find evidence, often on the web. In knowledgeable open dialogue~\cite{dinan2018wizard}, models need access to knowledge from large corpora to sustain informed conversations.

In general, solving \emph{knowledge-intensive} tasks requires--even for humans--access to a large body of information. Like in Information Retrieval (IR) this involves satisfying an information need leveraging large collections of text \cite{manning2008introduction}. However, while IR focuses of finding relevant material (usually documents), the tasks we consider focus on  more fine-grained behavior, such as producing specific answers to queries.
For such knowledge-intensive tasks, general infrastructure and architectures across tasks have yet to emerge, and fundamental research questions remain open. For example, while it was long assumed that non-parametric and explicit memory accessed through retrieval is strictly required for competitive results~\cite{chen2017reading}, recent large pre-trained sequence-to-sequence models such as T5~\cite{2019t5} and BART~\cite{Lewis2019BARTDS} store all knowledge in their parameters while performing remarkably well~\cite{petroni2019language}. 
Likewise, while the classical approach of information extraction for populating a Knowledge Base~\cite[KB, ][]{riedel2013relation,surdeanu2014overview}
seems out-of-fashion, recent results show that they remain contenders~\cite{fan-etal-2019-using,xiong2019pretrained}.

While there are numerous datasets for knowledge-intensive tasks~\citep[e.g.][to name just a few]{Thorne18Fever,dinan2018wizard,kwiatkowski2019natural}, it is difficult to answer the above questions generally across them. Each dataset comes in a different format, is pre-processed with different assumptions, 
and requires different loaders, evaluations, and analysis tools. 
Critically, they all use different knowledge sources, from different versions of Wikipedia to entirely different corpora. This makes task-to-task comparisons difficult and substantially increases computational overhead. For example, one cannot easily assess whether the same knowledge representation can be re-used if each dataset is tied to a different source. Moreover, if one decides to work with different sources across different tasks, many approaches require re-indexing and re-encoding large numbers of documents.
If a language model is pre-trained on one snapshot of Wikipedia to capture its knowledge, tasks that use other snapshots might require re-training.

To facilitate research on models that must access specific information in a knowledge source, we introduce \textbf{KILT}, a benchmark and library for \textbf{\emph{K}}nowledge~\textbf{\emph{I}}ntensive~\textbf{\emph{L}}anguage~\textbf{\emph{T}}asks. KILT aims to lower the entry barrier for such research by formulating several knowledge-intensive NLP tasks with respect to a common interface and the same unified knowledge source---a single Wikipedia snapshot. 
The KILT benchmark consists of eleven datasets spanning five distinct tasks, and includes the test set for all datasets considered.\footnote{A brand new portion of the Natural Question (NQ) dataset, originally held out, is used as the KILT test set for NQ.} 
An important aim of KILT is to cover many different ways of seeking knowledge. For this reason, we select tasks that provide a variety of ways to formulate both the input query (e.g., a claim to verify, a text chunk to annotate, a structured query, a natural question or a conversation) and the expected output (e.g., discrete, extractive, or abstractive). Moreover, while some tasks are factoid in nature (e.g., slot filling), others require using background knowledge to answer more complex questions (e.g, ELI5) or to sustain a conversation (e.g,. Wizard of Wikipedia). 
The format of the KILT benchmark is model-agnostic, so any system capable of producing a textual output given a textual input is eligible to participate. KILT is an \textit{in-KB} resource \cite{petroni2015core}, i.e., the evidence required to answer each of the \textasciitilde3.2M instances in KILT is present somewhere in the knowledge source. Hence there are no unanswerable instances in KILT. Although recognizing unanswerable instances is important, we believe the in-KB setting already poses an hard challenge to current state-of-the-art techniques, and thus leave unanswerable instances as future work.

KILT enables researchers to develop general-purpose models and evaluate them across multiple domains, testing hypotheses around task-agnostic memory and knowledge representations without 
%re-
indexing different large-scale textual corpora or writing new IO routines.
Furthermore, the KILT library provides general building blocks to ease research on knowledge intensive NLP. We provide various state-of-the-art information retrieval systems (both neural and non-neural) coupled with different models that read text in the knowledge source and make predictions for different tasks.
%Our interface for these components allows for easy interchange.\tim{interchange between what?} 

% This enables researchers to test new hypotheses, for example, around task-agnostic memory, without re-indexing new corpora, writing new IO routines, evaluation metrics etc. Kilt consists of 5 different tasks and 9 datasets, from fact checking to knowledge-intensive dialog. In total, \seb{Fabio, can you add some high level statistics here?} 
% \fabio{ single memory - benchmarking the best neural wikipedia . }

%\seb{Fill up results/outcome para}
% We got access to the official test set for all considered dataset, plus a brand new portion of the Natural Question (NQ) dataset \cite{kwiatkowski2019natural} (see the acknowledgments in Section \ref{acknowledgment}) that we use as official open-domain test set for NQ in KILT.

We evaluate several state-of-the-art models that represent diverse approaches to knowledge-intensive NLP, and find that a hybrid approach combining a neural retriever with a pretrained sequence-to-sequence model outperforms most task-specific solutions when trained end-to-end. % \fabio{especially when the signal is back-propagated through the two}. 
% Remarkably, such baseline achieves good performance on entity linking tasks thanks to a novel generative-based approach to the problem, getting numbers close to a dedicated state-of-the-art solution even if trained with less data.
We additionally evaluate whether systems can provide evidence for their predictions.
With this aim, we augment every instance in KILT with \textit{provenance} information in the form of textual spans in specific Wikipedia pages to corroborate the output. We additionally perform an annotation campaign via Amazon Mechanical Turk to increase the provenance coverage.
%for some datasets.\tim{vague: which datasets? why only some? what do we let annotators annotate?} 
Lastly, in addition to evaluating downstream performance with popular metrics we formulate novel KILT variants for those that award points only if systems find provenance Wikipedia pages for the output given the input.
The poor absolute performance of our baselines for those metrics indicates the need for focused research on systems able to explain their decisions.
%with improved provenance, to explain their decision.

In summary, we contribute:
\vspace{-0.55em}
\begin{enumerate} %[\itshape 1\upshape)]
\itemsep-0.45em 
\item a publicly-available benchmark of knowledge-intensive tasks aligned to a single Wikipedia snapshot, to spur the development of general-purpose models and enable their comparison;
\item an open-source library to facilitate the development of new architectures for knowledge-intensive tasks;
\item a provenance indication for all instances in KILT, made more comprehensive with an annotation campaign, which allows to jointly assess output accuracy and ability to provide supporting evidence in the knowledge source;
\item a comparative performance of various modeling approaches, showing promising results for general baselines across all tasks.
\end{enumerate}

\section{Knowledge Source}

%\fabio{section X, unified knowledge source in this section, we describe how we consolidate various nlp tasks that require knowledge into one fixed snapshot of wikipedia, which improves comparability and assessment on multiple tasks. generally, we assume tasks have the form: blah blah, such that they have provenance blah blah}
%\fabio{defines provenance/exact match, etc. then talk about how the mapping is done generally}

A main feature of the KILT benchmark is the use of a unified knowledge source that contains all information necessary for all tasks. Defining a unified knowledge source is a challenging problem --- although all tasks use Wikipedia, they consider different snapshots. As Wikipedia pages are constantly modified, added, and removed, the knowledge can differ drastically from snapshot to snapshot. 
Concretely, the KILT knowledge source is based on the 2019/08/01 Wikipedia snapshot and contains 5.9M articles. We describe how each dataset is represented in KILT, and our mapping strategy for aligning data to our chosen snapshot. Additional details are in the appendix.

\paragraph{Mapping Datasets to a Fixed Snapshot}
\label{sec:mapping}

The main challenge in defining a unified knowledge source is ensuring the knowledge for all task examples is available. We assume tasks provide an \emph{input} (e.g. a question in question answering, or a conversation in dialogue) needed to produce an \emph{output} (e.g. an answer or a subsequent utterance). In addition, tasks provide \emph{provenance}, defined as a set of textual spans in Wikipedia that contain evidence for producing an output given a specific input. These provenance spans range from single entities, short answers, sentences, paragraphs, to whole articles. 
% In particular, we collect \textit{provenance} information for each input-output pair, defined as a set of textual spans in Wikipedia that contain evidence for producing an output given a specific input.
The idea of our mapping strategy is to identify provenance spans in the KILT knowledge source---if we find all the provenance spans for an input-output pair, the knowledge needed to produce the output is available in our snapshot. The provenance can be a span of any size, from a single token to a paragraph to an entire document.

Concretely, the mapping strategy operates as follows.\footnote{Scripts for the mapping algorithm available on GitHub.} First, we try to match Wikipedia pages in each dataset to our snapshot, relying on Wikipedia URL redirections for pages that changed title. Second, we look for the provenance span in the matched page. We scan the whole page and return the span with the highest BLEU \cite{papineni2002bleu} with the given provenance span.\footnote{We return the shortest span if there's a tie in BLEU score.} Third, we replace the original provenance in a task's input-output pair with the span from the KILT knowledge source, and we report the BLEU score between the two. Finally, we remove from the dev and test sets all outputs for which the BLEU score is lower than a threshold for at least one provenance span (we use 0.5 as threshold) --- this is meant to ensure high quality mappings in the evaluation sets --- discarding on average 18\% of test and dev data (for all tasks except entity linking). We keep all input-output pairs in the train sets (see Figure \ref{fig:bleu} in the appendix for more details).

%\fabio{the threshould has been chosen for high-precision}

%\angela{we could give a sense of how much is filtered at each step? we start by saying we keep 99 percent, which somewhat raises the question of how much is kept in other steps}
%\patrick{I think we either should keep this bit purely a method description, and include concrete numbers later , or go more quantitative - this could be a good opportunity for a figure if we have the stats - a "Sankey diagram"? If someone can provide the numbers, i can make one - see fig \ref{fig:sankey_mockup} for a mockup of what this could look like
%}

% \iffalse
% \begin{figure*}[t!]
%     \centering
%     \includegraphics[width=0.8\linewidth]{figures/sankey_mockup}
%     \caption{KILT mapping efficiency \patrick{mockup! replace with real numbers, better aspect ratio, labels etc}}
%     \label{fig:sankey_mockup}
% \end{figure*}
% \fi

\section{Tasks}
\label{sec:tasks}

\begin{table*}[t!]
    \centering
\resizebox{\textwidth}{!}{    
    \fontsize{8.4}{10.1}\selectfont \setlength{\tabcolsep}{0.5em}
    \begin{tabular}{rllccc}
        \toprule
        \multicolumn{1}{c}{Label} & \multicolumn{1}{c}{Dataset} & \multicolumn{1}{c}{Reference} & Task & Input Format &  Output Format\\
        \midrule
        \textbf{FEV} &FEVER & \citet{Thorne18Fever} & Fact Checking & Claim & Classification \\
        \textbf{AY2} & AIDA CoNLL-YAGO & \citet{hoffart2011robust} &  Entity Linking  & Text Chunk & Entity \\
        \textbf{WnWi} & WNED-WIKI & \citet{guo2018robust} &  Entity Linking  & Text Chunk & Entity \\
        \textbf{WnCw} & WNED-CWEB & \citet{guo2018robust} &  Entity Linking  & Text Chunk & Entity \\
        \textbf{T-REx} & T-REx & \citet{elsahar2019t} & Slot Filling & Structured & Entity \\
        \textbf{zsRE} & Zero Shot RE & \citet{levy2017zero} &  Slot Filling & Structured & Entity \\
        \textbf{NQ} & Natural Questions & \citet{kwiatkowski2019natural} &  Open Domain QA & Question & Extractive \\
        \textbf{HoPo} & HotpotQA & \citet{yang2018hotpotqa} &  Open Domain QA & Question & Short Abstractive \\
        \textbf{TQA} & TriviaQA & \citet{joshi2017triviaqa} &  Open Domain QA  & Question & Extractive  \\
        \textbf{ELI5} & ELI5 & \citet{fan2019eli5} &  Open Domain QA  & Question & Long Abstractive \\
        \textbf{WoW} & Wizard of Wikipedia & \citet{dinan2018wizard}  & Dialogue & Conversation & Long Abstractive  \\
        \bottomrule
    \end{tabular}
}
    \caption{Datasets and tasks considered in KILT.}
    \label{tab:datasets}
\end{table*}

We consider five tasks that use Wikipedia as a knowledge source for KILT: fact checking, open domain question answering, slot filling, entity linking, and dialogue.  The diversity of these tasks challenge models to represent knowledge flexibly. Some tasks require a discrete prediction (e.g., an entity), others, such as extractive question answering, can copy the output directly from a Wikipedia page, while still other tasks must synthesize multiple pieces of knowledge in an abstractive way to produce an output. KILT also provides a variety of ways to seek knowledge, from a claim to verify to a text chunk to annotate, from a structured or natural question to a conversation (see Table \ref{tab:datasets} for details). 
We are able to include the test set for all datasets in KILT, either because the test set is public, or because we were able to obtain the test set from the authors of the original dataset. These test sets are not publicly released, but are used for the KILT challenge on EvalAI~\cite{EvalAI} 
where participants can upload their models' predictions and be listed on the public leaderboard.\footnote{available at \url{https://evalai.cloudcv.org/web/challenges/challenge-page/689}.}

To facilitate experimentation, we define a consistent interface for all datasets in the KILT Benchmark.
Each dataset is represented in JSON Line format
%\footnote{\url{http://jsonlines.org}}
, 
where each record contains three fields: \emph{id}, \emph{input}, \emph{output}. The \emph{input} is a natural language string and the \emph{output} a non-empty list of equally-valid outputs (e.g. if multiple answers to a question are valid in a question answering dataset). Each output is a string and it is accompanied by a non-empty list of complementary provenance spans (all should be used to acquire the knowledge needed to provide a valid output).
Figure \ref{fig:example} displays an example for all considered tasks (Figure \ref{fig:interface} in the appendix contains further details on the common interface). 

%\fabio{describe common interface}
%\fabio{example common interface for each task / figure}
%* Common interface: inputs, outputs:
%* Input sentence: question or claim
%* Output: a list of answers: each answer is string (label) and provenance (Wiki page)
%\fabio{then you do the datasts, and any details per dataset and discussed along with what the dataset is}
%\fabio{ok we do mapping, this is possible because dataset 1 has form X and datset 2 has form Y. but to deal with dataset3, we do strategy Z"  - maybe a bit more organized}

\subsection{Fact Checking}

Fact checking verifies a claim against a collection of evidence. It requires deep knowledge about the claim and reasoning over multiple documents. We consider the claim as \emph{input} and the classification label as \emph{output}. Each label is accompanied by a set of provenance spans that corroborate the classification label. We model multiple equally-valid provenance sets per label.

\textbf{FEVER}~\cite{Thorne18Fever} is a large dataset for claim veracity that requires retrieving sentence-level evidence to support if a claim is supported or refuted. Additional details are in the appendix.
% The task requires retrieving appropriate evidence at a sentence-level and using this to classify whether a claim is supported or refuted. 
%Optional sentence (can be cut if needed)
% The claims in the FEVER dataset are short factoid statements that are written and then mutated by annotators using sentences from the 50,000 most popular entities. 
%end optional sentence

%it is extremely difficult to assess if the knowledge is consistent (i.e., there is no evidence to assess the claim) in the Wikipedia snapshot we consider for KILT.  
%\jt{we could also argue (space permitting) perhaps that annotator recall is 72\% so some of these instances count be supported/refuted by evidence that was not found by annotators.}
%there is no evidence attached to them -  i.e. the right answer is that there is no evidence in the Wikipedia dump used in FEVER that completely supports/refutes them. 
%Unfortunately this makes

\subsection{Entity Linking}

Entity Linking (EL) assigns a unique Wikipedia page to entities mentioned in text. Each KILT record for EL has text in the \textit{input} (max 256 tokens) where a single entity mention is tagged with two special tokens (i.e., \textit{[START\_ENT]} and \textit{[END\_ENT]}---see Figure \ref{fig:example} for an example). The \textit{output} is the title of the Wikipedia page for the entity mention plus \textit{provenance} pointing to the entire page (through a unique identifier). Since Wikipedia associates unambiguous titles to entities\footnote{Wikipedia uses explicit text in titles to disambiguate.}, finding the correct output is enough to link entity mention and Wikipedia page. The \textit{provenance} mimics the canonical approach to EL, that is to produce an identifier for each mention~\cite{wu2019zero}.
To map the provenance (whole Wikipedia page), we simply match Wikipedia pages specified in  various datasets to the KILT knowledge source.  
We consider three popular EL datasets in KILT, two of which do not contain a train set but should be assessed in a zero-shot fashion. Note that, in addition to the AY2 train set, the whole knowledge source can be used as training data by exploiting hyperlinks. To facilitate experimentation, we release such data in KILT format (9M train instances), following the splits of~\citet{wu2019zero}.

\textbf{AIDA CoNLL-YAGO}~\cite{hoffart2011robust} supplements the CoNLL 2003 dataset~\cite{sang2003introduction} with Wikipedia URL annotations for all entities using the YAGO2 system \cite{hoffart2011yago2}. The original data is split into three parts: \textit{train}, \textit{testa}, \textit{testb}. Following \citet{hoffart2011robust} we consider \textit{testa} as dev and \textit{testb} as test.

\textbf{WNED-WIKI}~\cite{guo2018robust} is a dataset automatically created by sampling document from the 2013/06/06 Wikipedia dump, and balancing the difficulty of linking each mention (using a baseline as proxy). We randomly split the dataset into dev and test.

\textbf{WNED-CWEB}~\cite{guo2018robust} is a dataset created with the same strategy as   WNED-WIKI, but sampling from the ClueWeb 2012 corpora annotated with the FACC1 system.\footnote{\url{http://lemurproject.org/clueweb12}} Similarly, we randomly split into dev and test.

\subsection{Slot Filling}

The goal of the Slot Filling (SF) is to collect information on
certain relations (or slots) of entities (e.g., subject entity \textit{Albert Einstein} and relation \textit{educated\_at}) from large
collections of natural language texts. A potential application is structured Knowledge Base Population~\cite[KBP][]{surdeanu2014overview}. 
SF requires (1) disambiguation of the input entity and (2) acquiring relational knowledge for that entity. For KILT, we model the \textit{input} as a structured string \emph{subject entity \textit{[SEP]} relation}, the \textit{output} as a list of equally-valid object-entities, each one accompanied with \textit{provenance} where the subject-relation-object fact manifests.

\textbf{Zero Shot RE}~\cite{levy2017zero} is a dataset designed to translate relation extraction  
% (i.e., given a sentence, a subject entity and a relation, extract the object entity if any) 
into a reading comprehension problem.
We consider the open-domain version of this dataset and align the input/output with the KILT interface. Additional details are in the appendix.

\textbf{T-REx}~\cite{elsahar2019t} provides a large-scale collection of facts aligned to sentences in Wikipedia abstracts through distant supervision. We consider each sentence as \textit{provenance} and formulate the input as above (details in the appendix).

\subsection{Open Domain Question Answering}

Open domain Question Answering~\cite{chen2017reading} is the task of producing the correct answer for a question, without a predefined location for the answer. Standard tasks such as SQuAD~\cite{rajpurkar2016squad} provide an evidence document, but in open domain tasks, 
models must reason over an entire knowledge source (such as Wikipedia). We consider the question as \emph{input} and the answer as \emph{output} with dataset-specific \textit{provenance}.

% \citet{guu2020realm} consider this task ``one of the most knowledge-intensive tasks in natural language processing'' since, unlike traditional reading comprehension tasks (such as SQuAD \cite{rajpurkar2016squad}), no pre-identified document set (known to contain the answer) is provided in input, and models should reason over the whole knowledge source.

% \fabio{how does it fit in the common interface?}

\textbf{Natural Questions}~\cite{kwiatkowski2019natural} is a corpus of real questions issued to the Google search engine. Each question comes with an accompanied Wikipedia page with an annotated long answer (a paragraph) and a short answer (one or more entities).
%\patrick{mention our dedicated test set? I think we could emphasise that more?}
% In KILT we consider only the latter. 
We consider the open-version of the dataset and use both long and short answers spans as \textit{provenance}.
We collaborated with the authors of Natural Questions to access a held out, unpublished portion of the original dataset to form a new test set for KILT.
By construction each QA pair is associated with a single Wikipedia page, although other pages might contain enough evidence to answer the question. To increase the provenance coverage we perform an Amazon Mechanical Turk campaign for the dev and test sets and increase the average number of provenance pages per question from 1 to 1.57 (details in \cref{sec:annotation}).

\textbf{HotpotQA}~\cite{yang2018hotpotqa} requires multi-hop reasoning over multiple Wikipedia pages to answer each question. For each question-answer pair, a set of supporting sentences are provided, and we consider these as \textit{provenance}. We focus on the \textit{fullwiki} setting, where systems are required to retrieve and reason over the whole Wikipedia. 

\textbf{TriviaQA}~\cite{joshi2017triviaqa} is a collection of question-answer-evidence triples. Evidence documents are automatically gathered from Wikipedia or the Web. We consider only the Wikipedia case. 
We use the answer span as \textit{provenance} and consider the full version of the dev and test set. %, rather than the smaller subset of verified documents that are guaranteed to contain the answer.

\textbf{ELI5}~\cite{fan2019eli5}\footnote{\url{https://yjernite.github.io/lfqa.html}} is a collection of question-answer-evidence triples where the questions are complex, and the answers are long, explanatory, and free-form.  
For dev and test, we collect annotations using Amazon Mechanical Turk, asking evaluators to select which supporting documents from Wikipedia can be used to answer the question. We treat these as gold provenance annotations for evaluation (details in \cref{sec:annotation}).

\subsection{Dialogue}

Chitchat dialogue is the task of developing an engaging chatbot that can discuss a wide array of topics with a user, which often relies on topical, factual knowledge. For example, it would be difficult to have a conversation about ``grayhounds'' without any information about that dog breed. We consider the conversation history as \emph{input} and the next utterance as  \emph{output}.

\textbf{Wizard of Wikipedia}~\cite{dinan2018wizard} is a large dataset of conversation grounded with knowledge retrieved from Wikipedia. One speaker in the conversation must ground their utterances in a specific knowledge sentence, chosen from a Wikipedia page. The chosen sentence forms the \emph{provenance} for KILT.

\section{Provenance Annotation Campaign}
\label{sec:annotation}

We perform an Amazon Mechanical Turk campaign on the NQ and ELI5 datasets for the dev and test splits. While for the NQ our aim is to increase the provenance coverage (i.e., we already have a provenance page for each qa pair) for ELI5 we want to collect provenance information from scratch. For each question we ask annotators to indicate if four pre-determined passages contain enough evidence to answer the question and additionally highlight a salient span in them. We select the passages to annotate using our baseline retrieval models, namely Tf-idf, DPR, RAG and BLINK + flair (details in the Appendix).\footnote{for Tf-idf and BLINK + flair we consider the first passage in the retrieved page} We only consider passages with some tokens overlap with the gold answers (at least 10\%).

For NQ, we additionally include gold passages among those to annotate, with the twofold objective of controlling the quality of the annotation process and filter out questions that can't be answered given the KILT Wikipedia snapshot.\footnote{we present passages in random order to the annotator to exclude biases.} If no passage is selected by an annotator we ask to provide either another one from Wikipedia or an explanation. We collect three annotations for each passage, and insert the passage as new provenance for the question if at least two annotators found enough evidence to answer in it.   
The average inter-annotator agreement is 0.3 and 0.1 Cohen's kappa for NQ and ELI5 respectively. Note that ELI5 questions are in general more complex than NQ ones, the required answer is not an extracted span from a page but a free-form explanation that not always can be grounded in Wikipedia. 

To make ELI5 data more robust we computed the overlap between provenance passages and answers for each instance using ROUGE-L and manually annotate instances with low overlap (ROUGE-L < 0.15). Overall, we were able to collect provenance information for 1507 dev instances (3000 annotated) and 600 test instances (2000 annotated) for ELI5, with an average of 1.18 Wikipedia pages as provenance per instance. For NQ, we filter out on average 8\% of data (258 dev and 110 test instances) and include on average 1.57 Wikipedia pages as provenance per instance. Additional details in the Appendix, \cref{tab:datasetsstats}.
\section{Evaluation Metrics}

Various tasks in the KILT Benchmark need to be evaluated differently, which can make task-wide comparison challenging. 
Further, there are multiple aspects of each system that we want to assess, namely (1) downstream results, (2) performance in retrieving relevant evidence to corroborate a prediction and (3) a combination of the two.
We report different metrics to capture these aspects.\footnote{evaluation scripts available in GitHub.}

\paragraph{Downstream performance.}
We consider different metrics to capture the uniqueness of the different tasks in KILT and mimic the typical way to assess performance for each dataset. 
We use \textit{Accuracy} for tasks that require a discrete output (e.g., an entity); \textit{Exact Match} (EM) for tasks with extractive (i.e., Natural Questions, TriviaQA) or short abstractive output format (i.e., HotpotQA); finally, for tasks with long abstractive output format, we use \textit{ROUGE-L} \cite{lin2004rouge} for ELI5 and F1-score for Wizard of Wikipedia.
For EM and F1-score we follow standard post-processing to lowercase, strip articles, punctuation, and duplicate whitespace from gold and predicted output \cite{rajpurkar2016squad}. Note that Accuracy is equivalent to strict exact match, without post-processing. We report additional metrics for some datasets in the appendix (Table \ref{tab:FEV}-\ref{tab:WoW}).

\paragraph{Retrieval.}
We adopt a page-level formulation and measure the ability of a model to provide a set of Wikipedia pages as evidence for a prediction.\footnote{our evaluation scripts allow to evaluate retrieval performance at a more fine-grained level (e.g., paragraph).}
For most datasets in KILT a single page is enough to provide complete evidence, with the exception of FEVER (\textasciitilde12\% which requires more than one page) and HotpotQA (two pages are always required).
We consider the following retrieval metrics in KILT:
%\begin{itemize}
%\item 

\emph{R-precision}, calculated as $\frac{r}{R}$, where $R$ is the number of Wikipedia pages inside each provenance set and $r$ is the number of relevant pages among the top-$R$ retrieved pages. For most of the datasets $R=1$ and this formulation is equivalent to \emph{Precision@1}. Concretely, R-precision=1 if all Wikipedia pages in a provenance set are ranked at the top. We report the maximum value among all provenance sets for any given input.
%\item 

\emph{Recall@k}, calculated as $\frac{w}{n}$, where $n$ is the number of distinct provenance sets for a given input and $w$ is the number of complete provenance sets among the top-$k$ retrieved pages. For datasets that require more than one page of evidence (e.g., FEVER and HotpotQA), we use the lowest ranked page in each provenance set to determine its position and remove the other pages in the set from the rank.
%\end{itemize}
For both metrics, we report the mean over all test datapoints.

\begin{table}[t!]
    \centering
\resizebox{\linewidth}{!}{    
    \fontsize{8.4}{10.1}\selectfont \setlength{\tabcolsep}{0.5em}
    \begin{tabular}{lcc}
        \toprule
        Model &  \#Parameters\\
        \midrule
        % DrQA tf-idf~\cite{chen2017reading} & - \\
        Trans MemNet~\cite{dinan2018wizard} & 15.5M \\
        BERT (base)~\cite{devlin-etal-2019-bert} & 110M \\
        NSMN~\cite{nie2019combining}  & 199M +93M nt \\
        T5 (base)~\cite{raffel2019exploring} & 220M \\
        DPR~\cite{karpukhin2020dense}& 220M +15B idx \\
        BERT (large)~\cite{devlin-etal-2019-bert} & 340M \\
        BART (large)~\cite{Lewis2019BARTDS} & 406M \\
        RAG~\cite{lewis2020retrievalaugmented} & 626M +15B idx \\
        BLINK~\cite{wu2019zero} & 680M +6B idx \\
        %T5 (large)~\cite{raffel2019exploring} & 770M \\
        
        \bottomrule
    \end{tabular}
}
    \caption{Baselines considered and total number of their trainable parameters. Non trainable (nt) parameters and index (idx) sizes are also reported.}
    \label{tab:baselines}
\end{table}

\begin{table*}[ht]
\centering
\resizebox{\textwidth}{!}{   
\fontsize{8.4}{10.1}\selectfont \setlength{\tabcolsep}{0.5em}
\begin{tabular}{ll@{\hskip 2em}rrrrrrrrrrr}
 \toprule
   & \multicolumn{2}{r}{Fact Check.}  & \multicolumn{3}{c}{Entity Linking} & \multicolumn{2}{c}{Slot Filling} & \multicolumn{4}{c}{Open Domain QA} & \multicolumn{1}{c}{Dial.}  \\
 & model & \textbf{FEV}  & \textbf{AY2} & \textbf{WnWi} & \textbf{WnCw} & \textbf{T-REx} & \textbf{zsRE}  & \textbf{NQ} & \textbf{HoPo} & \textbf{TQA} & \textbf{ELI5} & \textbf{WoW}   \\
\midrule
 && \multicolumn{6}{c|}{Accuracy} & \multicolumn{3}{c|}{Exact Match} & \multicolumn{1}{c|}{RL} & \multicolumn{1}{c}{F1} \\
\midrule
\multirow{4}{*}{\rotatebox[origin=c]{90}{\textit{ts}}} & NSMN & 66.1 & - & - & - & - & - & - & - & - & - & -  \\
& BERT + DPR & 69.68 & - & - & - & - & 6.93 & 38.64 & 11.29 & 70.38 & - & -  \\
& BLINK  & - & \textbf{81.54} & \textbf{80.24} & \textbf{68.77} & - & - & - & - & -  & - & -  \\
& Trans MemNet & - & - & - & - & - & - & - & - & - & - & 11.85  \\
\cmidrule(lr){0-12}
\multirow{2}{*}{\rotatebox[origin=c]{90}{\textit{im}}} & BART (large) & 78.93 & 77.55 & 45.91 & 49.16 & 45.06 & 9.14 & 21.75 & 15.37 & 32.39 & \textbf{20.55} & 12.86 \\
& T5 (base) & 76.3 & 74.05 & 47.13 & 49.29 & 43.56 & 9.02 & 19.6 & 12.64 & 18.11 & 19.08 & 13.53\\
\cmidrule(lr){0-12}
\multirow{2}{*}{\rotatebox[origin=c]{90}{\textit{ex}}} & BART + DPR & \textbf{86.74} & 75.49 & 45.2 & 46.87 & 59.16 & 30.43 & 41.27 & 25.18 & 58.55 & 17.41 & \textbf{15.19} \\
& RAG & 86.31 & 72.62 & 48.07 & 47.61 & \textbf{59.2} & \textbf{44.74} & \textbf{44.39} & \textbf{26.97} & \textbf{71.27} & 14.05 & 13.11\\
 \bottomrule
\end{tabular}
}
\caption{Downstream performance on the test data. Baselines are grouped by task-specific (\textit{ts}) and general with implicit (\textit{im}) or explicit (\textit{ex}) knowledge access. Task-specific solutions cannot be generally applied to all datasets in KILT, hence there are empty cells in the top part of the table. We report the typical metric to assess performance for each dataset, specified in the first row.}
\label{tab:downstream_table}
\end{table*}

\paragraph{KILT scores.}
We propose a KILT version for downstream metrics that, inspired by the FEVER-score \cite{Thorne18Fever}, takes into account the provenance supporting the output.
For each datapoint, we only award Accuracy, EM, ROUGE-L, and F1 points to \textit{KILT-AC}, \textit{KILT-EM}, \textit{KILT-RL} and \textit{KILT-F1} respectively, if the R-precision is 1. This is equivalent to awarding points if the system finds (and ranks at the top) a complete set of provenance Wikipedia pages for at least one ground truth output given the input. We choose this metric to emphasize that systems must be able to explain their output with proper evidence, not simply answer.

\begin{table*}[t]
\centering

\resizebox{\textwidth}{!}{   

\fontsize{8.4}{10.1}\selectfont \setlength{\tabcolsep}{0.5em}
\begin{tabular}{l@{\hskip 3em}rrrrrrrrrrr}
 \toprule
% && \multicolumn{4}{l}{\textbf{~~CoarsexGrained}} & \multicolumn{6}{c}{\textbf{FinexGrained}}\\ % SB: Something odd is going on with centering. Fixing manually for now.
   \multicolumn{2}{r}{Fact Check.}  & \multicolumn{3}{c}{Entity Linking} & \multicolumn{2}{c}{Slot Filling} & \multicolumn{4}{c}{Open Domain QA} & \multicolumn{1}{c}{Dial.}  \\
 model  & \textbf{FEV}  & \textbf{AY2} & \textbf{WnWi} & \textbf{WnCw} & \textbf{T-REx} & \textbf{zsRE}  & \textbf{NQ} & \textbf{HoPo} & \textbf{TQA} & \textbf{ELI5} & \textbf{WoW}   \\
\midrule
& \multicolumn{11}{c}{R-Precision} \\
\midrule
DPR + BERT  & {72.93} & - & - & - & - & 40.11 & \textbf{60.66} & 25.04 & 43.4 & - & -\\
DPR  & 55.33 & 1.81 & 0.3 & 0.51 & 13.26 & 28.96 & 54.29 & 25.04 & 44.49 & 10.67 & 25.48 \\ 
 Multi-task DPR & \textbf{74.48} & 26.48 & 4.91 & 1.86 & \textbf{69.46} & \textbf{80.91} & 59.42 & 42.92 & 61.49 & \textbf{15.5} & 41.07 \\
Tf-idf  & 50.85 & 3.74 & 0.24 & 2.09 & 44.74 & 60.83 & 28.12 & 34.14 & 46.37 & {13.67} & 49.01 \\ 
RAG  & 61.94 & 72.62 & 48.07 & 47.61 & 28.68 & 53.73 & 59.49 & 30.59 & 48.68 & 11.0 & \textbf{57.78} \\
BLINK + flair & 63.71 & \textbf{81.54} & \textbf{80.24} & \textbf{68.77} & {59.56} & {78.78} & 24.52 & \textbf{46.12} & \textbf{65.58}  & 9.5 & 38.21 \\
\bottomrule
\end{tabular}
}
\caption{Page-level R-Precision on test data. For DPR, we additionally report the performance after the BERT-based classifier (for FE) or reader (for NQ,HP,TR) re-ranked relevant pages (i.e., DPR + BERT). R-Precision is equivalent to Precision@1 for all datasets except FEV and HoPo that require multi-hop.}
\label{tab:test-retrieval-results}
\end{table*}
\begin{table*}[ht]
\centering
\resizebox{\textwidth}{!}{   
\fontsize{8.4}{10.1}\selectfont \setlength{\tabcolsep}{0.5em}
\begin{tabular}{ll@{\hskip 2em}rrrrrrrrrrr}
 \toprule
   & \multicolumn{2}{r}{Fact Check.}  & \multicolumn{3}{c}{Entity Linking} & \multicolumn{2}{c}{Slot Filling} & \multicolumn{4}{c}{Open Domain QA} & \multicolumn{1}{c}{Dial.}  \\
 & model & \textbf{FEV}  & \textbf{AY2} & \textbf{WnWi} & \textbf{WnCw} & \textbf{T-REx} & \textbf{zsRE}  & \textbf{NQ} & \textbf{HoPo} & \textbf{TQA} & \textbf{ELI5} & \textbf{WoW}   \\
\midrule
 && \multicolumn{6}{c|}{KILT-AC} & \multicolumn{3}{c|}{KILT-EM} & \multicolumn{1}{c|}{-RL} & \multicolumn{1}{c}{-F1} \\
\midrule
\multirow{4}{*}{\rotatebox[origin=c]{90}{\textit{ts}}} & NSMN & 41.88 & - & - & - & - & - & - & - & - & - & -  \\
& BERT + DPR & \textbf{58.58}  & -  & - & - & - & 4.47 & 31.99 & 0.74 & 34.48 & - & - \\
& BLINK  & - & \textbf{81.54} & \textbf{80.24} & \textbf{68.77} & - & - & - & - & - & - & -  \\
& Trans MemNet & - & - & - & - & - & - & - & - & - & - &  2.2  \\
\cmidrule(lr){0-12}
\multirow{2}{*}{\rotatebox[origin=c]{90}{\textit{ex}}} & BART + DPR & 47.68 & 75.49 & 45.2 & 46.87 & 11.12 & 18.91 & 30.06 & 1.96 & 31.4 & \textbf{1.9} & 4.37 \\
& RAG & 53.45 & 72.62 & 48.07 & 47.61 & \textbf{23.12} & \textbf{36.83} & \textbf{32.69} & \textbf{3.21} & \textbf{38.13} & 1.69 & \textbf{8.75} \\
 \bottomrule
\end{tabular}
}
\caption{
KILT scores  on the test data.
We do not report KILT scores for baselines with implicit knowledge access since no provenance information is returned by them. We report the KILT version of donwstream metrics, specified in the first row (to save space we abbreviate KILT-RL and KILT-F1). KILT scores are computed by awarding points only if provenance pages are found (i.e., R-Precision = 1). }
\label{tab:KILT_table}
\end{table*}

\section{Baselines}

The KILT tasks provide a dual challenge of retrieving information and conditioning upon that to create an output. Various directions could be applied to these. For example, the Wikipedia knowledge could be represented \emph{explicitly}, as natural language or in a structured form, or represented \emph{implicitly}, as knowledge stored in model parameters. Models could be \textit{discriminative}, \emph{extractive}, where a specific span is selected as output, or \emph{generative}, where the model writes an output.
We consider retrieval, task-specific, and general baselines for KILT (see Table \ref{tab:baselines}). Additional details are in the appendix.
\section{Results}

%\fabio{main 2-3 points we can go for? e.g. "a lot of approaches are promising, X is working the best, and we'll release these models and have them available in KILT library so everyone can improve on these}

We summarize the main results in three tables: downstream performance in \Cref{tab:downstream_table}, retrieval in \Cref{tab:test-retrieval-results} and KILT scores in \Cref{tab:KILT_table}. Additional results, as well as comparisons with recent works reported numbers, can be found in the appendix. It's possible to get the performance of a system for the KILT test sets by uploading its predictions to our EvalAI challenge.\footnotemark[5]

%\paragraph{General vs Task-specific solutions.} 
When considering downstream performance (\Cref{tab:downstream_table}), although pre-trained sequence-to-sequence models can embed knowledge implicitly in their parameters to some extent~\cite{petroni2019language,roberts2020much}, they clearly lag behind models with explicit knowledge access in almost all datasets.
The BART+DPR baseline that incorporates an explicit retrieval step in addition to the generative pretraining, works well. It outperforms some of the task-specific solutions, and gets close to others. Performance are even stronger when the retriever and reader components are trained end-to-end, as in the case of RAG. We find this a promising direction for knowledge intensive tasks.

By formulating Entity Linking within KILT, we can evaluate the ability of seq2seq models at this task. They perform surprisingly well, even without any explicit access to knowledge (i.e., BART and T5). These solutions are able to link entity mentions by either leaving them untouched (if they match the correct Wikipedia title), completely altering mention text (e.g.,  ``European Cup'' 	$\rightarrow$ ``UEFA Champions League''), or adding disambiguation tokens (e.g., ``Galatasaray''	$\rightarrow$ ``Galatasaray S.K. (football)''). 
We report an example in \Cref{fig:entitylinking}. 

When considering retrieval alone (Table \ref{tab:test-retrieval-results}) there is no clear winner---entity-centric tasks (Entity Linking and Slot Filling) clearly benefit from entity-based retrieval, while DPR works better for NQ, FEV and ELI5, that require more fine grained passages supervision. We believe that combining all these ingredients (i.e., dense representations, fine grained supervision, entity awareness) will be necessary for general task-agnostic memories.
Moreover, jointly training a single DPR model on all KILT training data (Multi-task DPR) led to strong performance gains on all datasets compared with the original model (DPR), that considers only NQ and TQA as training data~\cite{karpukhin2020dense}. This suggests synergies between KILT datasets that are beneficial in terms of model performance.

%Note that although F1-score is not a standard way to report EL results, KILT-F1-score is. With the latter, we award points only if the correct Wikipedia page is identified as provenance. 
  
%that work well across all tasks in KILT.

Finally, the KILT scores formulation allows us to systematically assesses the performance for output and provenance jointly (\Cref{tab:KILT_table}).
%---each system needs to be able to provide the correct output but also successfully justify its decision. 
We don't report results for BART and T5 since answers are generated solely from the input with no explicit retrieval and there is no straightforward way to access provenance for each prediction.
The relative performance of the other baselines with respect to KILT scores is consistent with downstream results. However, the generally low absolute numbers leave a large room for improvement for systems able to provide the correct output but also successfully justify their decision.
\section{Discussion}

%\fabio{metric discussion}

%\fabio{the difficulty of defining a single score, around leaderboards in general?}
There are custom solutions that can easily simplify the slot filling task. For instance, subject entities can be used for lookups by title in Wikipedia to retrieve knowledge (this heuristic will always work for zsRE), and structured human-curated resources (such as Wikidata\footnote{\url{https://www.wikidata.org}}) could be used to get all answers right. 
Nevertheless, we are interested in testing if a general model can extract attributes about specific entities from a large body of text.

%by manipulating the Zero Shot RE dataset. However, there are custom solutions that can easily simplify the task. For instance, subject entities can be used for lookups by title in Wikipedia to retrieve knowledge, and structured human-curated resources (such as Wikidata\footnote{\url{https://www.wikidata.org}}) could be used to get all answers right. 
%Nevertheless, we are interested in testing if a general model can extract attributes about specific entities from a large body of text. 
% Hence, we consider ZS as a sort of ``unit test'' for KILT.

The provenance to justify each system prediction can come from anywhere, including a different system, and this is difficult to detect.
Moreover our provenance might not be exhaustive---given the redundancy of information in Wikipedia there could be other pages with the knowledge needed to solve a KILT instance. We conduct an annotation campaign to mitigate the problem.

\section{Related Work}

%\fabio{convincing people to condition to a large external textual corpora}

%\fabio{it can be sort}

%\begin{itemize}
%    \item~GLUE \cite{wang2018glue}
%    \item~super GLUE \cite{wang2019superglue}
%    \item~decaNLP \cite{mccann2018natural}
%    \item~The Dialogue Dodecathlon \cite{shuster2019dialogue} 
%    \item~ORB \cite{dua2019orb}
%    \item~BREAK \cite{wolfson2020break}
%\end{itemize}
    
Several natural language benchmarks have been introduced to track and support NLP progress, including natural language understanding~\cite{wang2018glue,wang2019superglue}, multitask question answering~\cite{mccann2018natural}, reading comprehension~\cite{dua2019orb}, question understanding~\cite{wolfson2020break}, and dialogue~\cite{shuster2019dialogue}. We focus on multi-domain tasks that need to seek knowledge in a large body of documents to produce an output.
Although there exist several tasks and resources that define large-scale external knowledge sources---including the TAC-KBP challenges~\cite{mcnamee2009overview, ji2010overview, surdeanu2013overview, surdeanu2014overview}, ARC~\cite{clark2018think}, TriviaQA-web~\cite{joshi2017triviaqa}, Quasar-T~\cite{dhingra2017quasar}, WebQuestions~\cite{berant2013semantic} and ComplexWebQuestions~\cite{talmor2018web}---in KILT we exclusively consider publicly available Wikipedia-based datasets in order to merge and unify the knowledge source.
\section{Conclusion}

We introduce KILT, a benchmark for assessing models that need to condition on specific  knowledge in a defined snapshot of Wikipedia to solve tasks spanning five domains. The goal is to catalyze and facilitate research towards general and explainable models equipped with task-agnostic  representations of knowledge.
Our experiments show promising results for a general solution combining dense retrieval and seq2seq generations, although there is large room for improvements.
In particular, we find that provenance of current models is generally low.
% Finally, we plan to explore multi-task learning to exploit synergies between KILT tasks and datasets in the future, and to develop general approaches for representing large-scale textual knowledge sources that are useful for multiple downstream tasks.
% \fabio{Future work: leverage synergies between this tasks for improve performance when training in a multi-task fashion.}

% future work
%\fabio{structured zero shot}

% TO REINTRODUCE FOR CR
\section{Acknowledgment}
\label{acknowledgment}

The authors would like to greatly thank the team behind Natural Questions\footnote{\url{https://ai.google.com/research/NaturalQuestions}} for the held out data, that defines our NQ test set;  FEVER\footnote{\url{https://fever.ai}}, HotpotQA\footnote{\url{https://hotpotqa.github.io}} and TriviaQA\footnote{\url{https://nlp.cs.washington.edu/triviaqa}} teams for sharing official test data for the KILT leaderboard; Luke Zettlemoyer and Scott Wen-tau Yih for helpful discussions; Rishabh Jain for the help in setting up the EvalAI challenge.

\bibliography{KILT}

\begin{thebibliography}{59}
\expandafter\ifx\csname natexlab\endcsname\relax\def\natexlab#1{#1}\fi

\bibitem[{Akbik et~al.(2019)Akbik, Bergmann, Blythe, Rasul, Schweter, and
  Vollgraf}]{akbik2019flair}
Alan Akbik, Tanja Bergmann, Duncan Blythe, Kashif Rasul, Stefan Schweter, and
  Roland Vollgraf. 2019.
\newblock Flair: An easy-to-use framework for state-of-the-art nlp.
\newblock In \emph{Proceedings of the 2019 Conference of the North American
  Chapter of the Association for Computational Linguistics (Demonstrations)},
  pages 54--59.

\bibitem[{Berant et~al.(2013)Berant, Chou, Frostig, and
  Liang}]{berant2013semantic}
Jonathan Berant, Andrew Chou, Roy Frostig, and Percy Liang. 2013.
\newblock Semantic parsing on freebase from question-answer pairs.
\newblock In \emph{Proceedings of the 2013 conference on empirical methods in
  natural language processing}, pages 1533--1544.

\bibitem[{Chen et~al.(2017)Chen, Fisch, Weston, and Bordes}]{chen2017reading}
Danqi Chen, Adam Fisch, Jason Weston, and Antoine Bordes. 2017.
\newblock Reading wikipedia to answer open-domain questions.
\newblock \emph{ACL}.

\bibitem[{Clark et~al.(2018)Clark, Cowhey, Etzioni, Khot, Sabharwal, Schoenick,
  and Tafjord}]{clark2018think}
Peter Clark, Isaac Cowhey, Oren Etzioni, Tushar Khot, Ashish Sabharwal, Carissa
  Schoenick, and Oyvind Tafjord. 2018.
\newblock Think you have solved question answering? try arc, the ai2 reasoning
  challenge.
\newblock \emph{arXiv preprint arXiv:1803.05457}.

\bibitem[{Devlin et~al.(2019)Devlin, Chang, Lee, and
  Toutanova}]{devlin-etal-2019-bert}
Jacob Devlin, Ming-Wei Chang, Kenton Lee, and Kristina Toutanova. 2019.
\newblock \href {https://doi.org/10.18653/v1/N19-1423} {{BERT}: Pre-training of
  deep bidirectional transformers for language understanding}.
\newblock In \emph{Proceedings of the 2019 Conference of the North {A}merican
  Chapter of the Association for Computational Linguistics: Human Language
  Technologies, Volume 1 (Long and Short Papers)}, pages 4171--4186,
  Minneapolis, Minnesota. Association for Computational Linguistics.

\bibitem[{Dhingra et~al.(2017)Dhingra, Mazaitis, and Cohen}]{dhingra2017quasar}
Bhuwan Dhingra, Kathryn Mazaitis, and William~W Cohen. 2017.
\newblock Quasar: Datasets for question answering by search and reading.
\newblock \emph{arXiv preprint arXiv:1707.03904}.

\bibitem[{Dinan et~al.(2019)Dinan, Roller, Shuster, Fan, Auli, and
  Weston}]{dinan2018wizard}
Emily Dinan, Stephen Roller, Kurt Shuster, Angela Fan, Michael Auli, and Jason
  Weston. 2019.
\newblock Wizard of wikipedia: Knowledge-powered conversational agents.
\newblock \emph{Proceedings of the International Conference on Learning
  Representations (ICLR)}.

\bibitem[{Dua et~al.(2019)Dua, Gottumukkala, Talmor, Singh, and
  Gardner}]{dua2019orb}
Dheeru Dua, Ananth Gottumukkala, Alon Talmor, Sameer Singh, and Matt Gardner.
  2019.
\newblock Orb: An open reading benchmark for comprehensive evaluation of
  machine reading comprehension.
\newblock \emph{arXiv preprint arXiv:1912.12598}.

\bibitem[{Elsahar et~al.(2018)Elsahar, Vougiouklis, Remaci, Gravier, Hare,
  Simperl, and Laforest}]{elsahar2019t}
Hady Elsahar, Pavlos Vougiouklis, Arslen Remaci, Christophe Gravier, Jonathon
  Hare, Elena Simperl, and Frederique Laforest. 2018.
\newblock T-rex: A large scale alignment of natural language with knowledge
  base triples.
\newblock \emph{LREC}.

\bibitem[{Fan et~al.(2019{\natexlab{a}})Fan, Gardent, Braud, and
  Bordes}]{fan-etal-2019-using}
Angela Fan, Claire Gardent, Chlo{\'e} Braud, and Antoine Bordes.
  2019{\natexlab{a}}.
\newblock \href {https://doi.org/10.18653/v1/D19-1428} {Using local knowledge
  graph construction to scale {S}eq2{S}eq models to multi-document inputs}.
\newblock In \emph{Proceedings of the 2019 Conference on Empirical Methods in
  Natural Language Processing and the 9th International Joint Conference on
  Natural Language Processing (EMNLP-IJCNLP)}, pages 4186--4196, Hong Kong,
  China. Association for Computational Linguistics.

\bibitem[{Fan et~al.(2019{\natexlab{b}})Fan, Jernite, Perez, Grangier, Weston,
  and Auli}]{fan2019eli5}
Angela Fan, Yacine Jernite, Ethan Perez, David Grangier, Jason Weston, and
  Michael Auli. 2019{\natexlab{b}}.
\newblock \href {https://doi.org/10.18653/v1/p19-1346} {{ELI5:} long form
  question answering}.
\newblock In \emph{Proceedings of the 57th Conference of the Association for
  Computational Linguistics, {ACL} 2019, Florence, Italy, July 28- August 2,
  2019, Volume 1: Long Papers}, pages 3558--3567. Association for Computational
  Linguistics.

\bibitem[{Ferragina and Scaiella(2011)}]{ferragina2011fast}
Paolo Ferragina and Ugo Scaiella. 2011.
\newblock Fast and accurate annotation of short texts with wikipedia pages.
\newblock \emph{IEEE software}, 29(1):70--75.

\bibitem[{Guo and Barbosa(2018)}]{guo2018robust}
Zhaochen Guo and Denilson Barbosa. 2018.
\newblock Robust named entity disambiguation with random walks.
\newblock \emph{Semantic Web}, 9(4):459--479.

\bibitem[{Guu et~al.(2020)Guu, Lee, Tung, Pasupat, and Chang}]{guu2020realm}
Kelvin Guu, Kenton Lee, Zora Tung, Panupong Pasupat, and Ming-Wei Chang. 2020.
\newblock \href {http://arxiv.org/abs/2002.08909} {Realm: Retrieval-augmented
  language model pre-training}.

\bibitem[{Hoffart et~al.(2011{\natexlab{a}})Hoffart, Suchanek, Berberich,
  Lewis-Kelham, De~Melo, and Weikum}]{hoffart2011yago2}
Johannes Hoffart, Fabian~M Suchanek, Klaus Berberich, Edwin Lewis-Kelham,
  Gerard De~Melo, and Gerhard Weikum. 2011{\natexlab{a}}.
\newblock Yago2: exploring and querying world knowledge in time, space,
  context, and many languages.
\newblock In \emph{Proceedings of the 20th international conference companion
  on World wide web}, pages 229--232.

\bibitem[{Hoffart et~al.(2011{\natexlab{b}})Hoffart, Yosef, Bordino,
  F{\"u}rstenau, Pinkal, Spaniol, Taneva, Thater, and
  Weikum}]{hoffart2011robust}
Johannes Hoffart, Mohamed~Amir Yosef, Ilaria Bordino, Hagen F{\"u}rstenau,
  Manfred Pinkal, Marc Spaniol, Bilyana Taneva, Stefan Thater, and Gerhard
  Weikum. 2011{\natexlab{b}}.
\newblock Robust disambiguation of named entities in text.
\newblock In \emph{Proceedings of the Conference on Empirical Methods in
  Natural Language Processing}, pages 782--792. Association for Computational
  Linguistics.

\bibitem[{Ji et~al.(2010)Ji, Grishman, Dang, Griffitt, and
  Ellis}]{ji2010overview}
Heng Ji, Ralph Grishman, Hoa~Trang Dang, Kira Griffitt, and Joe Ellis. 2010.
\newblock Overview of the tac 2010 knowledge base population track.
\newblock In \emph{Third text analysis conference (TAC 2010)}, volume~3, pages
  3--3.

\bibitem[{Joshi et~al.(2017)Joshi, Choi, Weld, and
  Zettlemoyer}]{joshi2017triviaqa}
Mandar Joshi, Eunsol Choi, Daniel Weld, and Luke Zettlemoyer. 2017.
\newblock \href {https://doi.org/10.18653/v1/P17-1147} {{T}rivia{QA}: A large
  scale distantly supervised challenge dataset for reading comprehension}.
\newblock In \emph{Proceedings of the 55th Annual Meeting of the Association
  for Computational Linguistics (Volume 1: Long Papers)}, pages 1601--1611,
  Vancouver, Canada. Association for Computational Linguistics.

\bibitem[{Karpukhin et~al.(2020)Karpukhin, O{\u{g}}uz, Min, Wu, Edunov, Chen,
  and Yih}]{karpukhin2020dense}
Vladimir Karpukhin, Barlas O{\u{g}}uz, Sewon Min, Ledell Wu, Sergey Edunov,
  Danqi Chen, and Wen-tau Yih. 2020.
\newblock Dense passage retrieval for open-domain question answering.
\newblock \emph{arXiv preprint arXiv:2004.04906}.

\bibitem[{Kwiatkowski et~al.(2019)Kwiatkowski, Palomaki, Redfield, Collins,
  Parikh, Alberti, Epstein, Polosukhin, Devlin, Lee
  et~al.}]{kwiatkowski2019natural}
Tom Kwiatkowski, Jennimaria Palomaki, Olivia Redfield, Michael Collins, Ankur
  Parikh, Chris Alberti, Danielle Epstein, Illia Polosukhin, Jacob Devlin,
  Kenton Lee, et~al. 2019.
\newblock Natural questions: a benchmark for question answering research.
\newblock \emph{Transactions of the Association for Computational Linguistics},
  7:453--466.

\bibitem[{Lee et~al.(2019)Lee, Chang, and Toutanova}]{lee2019latent}
Kenton Lee, Ming-Wei Chang, and Kristina Toutanova. 2019.
\newblock Latent retrieval for weakly supervised open domain question
  answering.
\newblock \emph{arXiv preprint arXiv:1906.00300}.

\bibitem[{Levy et~al.(2017)Levy, Seo, Choi, and Zettlemoyer}]{levy2017zero}
Omer Levy, Minjoon Seo, Eunsol Choi, and Luke Zettlemoyer. 2017.
\newblock Zero-shot relation extraction via reading comprehension.
\newblock \emph{CoNLL}.

\bibitem[{Lewis et~al.(2020{\natexlab{a}})Lewis, Ghazvininejad, Ghosh,
  Aghajanyan, Wang, and Zettlemoyer}]{lewis2020pre}
Mike Lewis, Marjan Ghazvininejad, Gargi Ghosh, Armen Aghajanyan, Sida Wang, and
  Luke Zettlemoyer. 2020{\natexlab{a}}.
\newblock Pre-training via paraphrasing.
\newblock \emph{arXiv preprint arXiv:2006.15020}.

\bibitem[{Lewis et~al.(2019)Lewis, Liu, Goyal, Ghazvininejad, Mohamed, Levy,
  Stoyanov, and Zettlemoyer}]{Lewis2019BARTDS}
Mike Lewis, Yinhan Liu, Naman Goyal, Marjan Ghazvininejad, Abdelrahman Mohamed,
  Omer Levy, Ves Stoyanov, and Luke Zettlemoyer. 2019.
\newblock Bart: Denoising sequence-to-sequence pre-training for natural
  language generation, translation, and comprehension.
\newblock \emph{ArXiv}, abs/1910.13461.

\bibitem[{Lewis et~al.(2020{\natexlab{b}})Lewis, Perez, Piktus, Petroni,
  Karpukhin, Goyal, K{\"u}ttler, Lewis, tau Yih, Rockt{\"a}schel, Riedel, and
  Kiela}]{lewis2020retrievalaugmented}
Patrick Lewis, Ethan Perez, Aleksandara Piktus, Fabio Petroni, Vladimir
  Karpukhin, Naman Goyal, Heinrich K{\"u}ttler, Mike Lewis, Wen tau Yih, Tim
  Rockt{\"a}schel, Sebastian Riedel, and Douwe Kiela. 2020{\natexlab{b}}.
\newblock \href {http://arxiv.org/abs/2005.11401} {Retrieval-augmented
  generation for knowledge-intensive nlp tasks}.

\bibitem[{Lin(2004)}]{lin2004rouge}
Chin-Yew Lin. 2004.
\newblock Rouge: A package for automatic evaluation of summaries.
\newblock In \emph{Text summarization branches out}, pages 74--81.

\bibitem[{Liu et~al.(2019)Liu, Ott, Goyal, Du, Joshi, Chen, Levy, Lewis,
  Zettlemoyer, and Stoyanov}]{liu2019roberta}
Yinhan Liu, Myle Ott, Naman Goyal, Jingfei Du, Mandar Joshi, Danqi Chen, Omer
  Levy, Mike Lewis, Luke Zettlemoyer, and Veselin Stoyanov. 2019.
\newblock Roberta: A robustly optimized bert pretraining approach.
\newblock \emph{arXiv preprint arXiv:1907.11692}.

\bibitem[{Manning et~al.(2008)Manning, Sch{\"u}tze, and
  Raghavan}]{manning2008introduction}
Christopher~D Manning, Hinrich Sch{\"u}tze, and Prabhakar Raghavan. 2008.
\newblock \emph{Introduction to information retrieval}.
\newblock Cambridge university press.

\bibitem[{McCann et~al.(2018)McCann, Keskar, Xiong, and
  Socher}]{mccann2018natural}
Bryan McCann, Nitish~Shirish Keskar, Caiming Xiong, and Richard Socher. 2018.
\newblock The natural language decathlon: Multitask learning as question
  answering.
\newblock \emph{arXiv preprint arXiv:1806.08730}.

\bibitem[{McNamee and Dang(2009)}]{mcnamee2009overview}
Paul McNamee and Hoa~Trang Dang. 2009.
\newblock Overview of the tac 2009 knowledge base population track.
\newblock In \emph{Text Analysis Conference (TAC)}, volume~17, pages 111--113.
  National Institute of Standards and Technology (NIST) Gaithersburg,
  Maryland~{\ldots}.

\bibitem[{Miller et~al.(2017)Miller, Feng, Fisch, Lu, Batra, Bordes, Parikh,
  and Weston}]{miller2017parlai}
Alexander~H Miller, Will Feng, Adam Fisch, Jiasen Lu, Dhruv Batra, Antoine
  Bordes, Devi Parikh, and Jason Weston. 2017.
\newblock Parlai: A dialog research software platform.
\newblock \emph{arXiv preprint arXiv:1705.06476}.

\bibitem[{Nie et~al.(2019)Nie, Chen, and Bansal}]{nie2019combining}
Yixin Nie, Haonan Chen, and Mohit Bansal. 2019.
\newblock Combining fact extraction and verification with neural semantic
  matching networks.
\newblock In \emph{Proceedings of the AAAI Conference on Artificial
  Intelligence}, volume~33, pages 6859--6866.

\bibitem[{Ott et~al.(2019)Ott, Edunov, Baevski, Fan, Gross, Ng, Grangier, and
  Auli}]{ott2019fairseq}
Myle Ott, Sergey Edunov, Alexei Baevski, Angela Fan, Sam Gross, Nathan Ng,
  David Grangier, and Michael Auli. 2019.
\newblock fairseq: A fast, extensible toolkit for sequence modeling.
\newblock In \emph{Proceedings of the 2019 Conference of the North American
  Chapter of the Association for Computational Linguistics (Demonstrations)},
  pages 48--53.

\bibitem[{Papineni et~al.(2002)Papineni, Roukos, Ward, and
  Zhu}]{papineni2002bleu}
Kishore Papineni, Salim Roukos, Todd Ward, and Wei-Jing Zhu. 2002.
\newblock Bleu: a method for automatic evaluation of machine translation.
\newblock In \emph{Proceedings of the 40th annual meeting on association for
  computational linguistics}, pages 311--318. Association for Computational
  Linguistics.

\bibitem[{Petroni et~al.(2015)Petroni, Corro, and Gemulla}]{petroni2015core}
Fabio Petroni, Luciano~del Corro, and Rainer Gemulla. 2015.
\newblock Core: Context-aware open relation extraction with factorization
  machines.
\newblock In \emph{EMNLP}. Assoc. for Computational Linguistics.

\bibitem[{Petroni et~al.(2020)Petroni, Lewis, Piktus, Rockt{\"a}schel, Wu,
  Miller, and Riedel}]{petroni2020context}
Fabio Petroni, Patrick Lewis, Aleksandra Piktus, Tim Rockt{\"a}schel, Yuxiang
  Wu, Alexander~H Miller, and Sebastian Riedel. 2020.
\newblock How context affects language models' factual predictions.
\newblock \emph{AKBC}.

\bibitem[{Petroni et~al.(2019)Petroni, Rockt{\"a}schel, Lewis, Bakhtin, Wu,
  Miller, and Riedel}]{petroni2019language}
Fabio Petroni, Tim Rockt{\"a}schel, Patrick Lewis, Anton Bakhtin, Yuxiang Wu,
  Alexander~H Miller, and Sebastian Riedel. 2019.
\newblock Language models as knowledge bases?
\newblock \emph{EMNLP}.

\bibitem[{Raffel et~al.(2019{\natexlab{a}})Raffel, Shazeer, Roberts, Lee,
  Narang, Matena, Zhou, Li, and Liu}]{2019t5}
Colin Raffel, Noam Shazeer, Adam Roberts, Katherine Lee, Sharan Narang, Michael
  Matena, Yanqi Zhou, Wei Li, and Peter~J. Liu. 2019{\natexlab{a}}.
\newblock \href {http://arxiv.org/abs/1910.10683} {Exploring the limits of
  transfer learning with a unified text-to-text transformer}.
\newblock \emph{arXiv e-prints}.

\bibitem[{Raffel et~al.(2019{\natexlab{b}})Raffel, Shazeer, Roberts, Lee,
  Narang, Matena, Zhou, Li, and Liu}]{raffel2019exploring}
Colin Raffel, Noam Shazeer, Adam Roberts, Katherine Lee, Sharan Narang, Michael
  Matena, Yanqi Zhou, Wei Li, and Peter~J Liu. 2019{\natexlab{b}}.
\newblock Exploring the limits of transfer learning with a unified text-to-text
  transformer.
\newblock \emph{arXiv preprint arXiv:1910.10683}.

\bibitem[{Rajpurkar et~al.(2016)Rajpurkar, Zhang, Lopyrev, and
  Liang}]{rajpurkar2016squad}
Pranav Rajpurkar, Jian Zhang, Konstantin Lopyrev, and Percy Liang. 2016.
\newblock Squad: 100,000+ questions for machine comprehension of text.
\newblock \emph{EMNLP}.

\bibitem[{Riedel et~al.(2013)Riedel, Yao, McCallum, and
  Marlin}]{riedel2013relation}
Sebastian Riedel, Limin Yao, Andrew McCallum, and Benjamin~M Marlin. 2013.
\newblock Relation extraction with matrix factorization and universal schemas.
\newblock In \emph{Proceedings of the 2013 Conference of the North American
  Chapter of the Association for Computational Linguistics: Human Language
  Technologies}, pages 74--84.

\bibitem[{Roberts et~al.(2020)Roberts, Raffel, and Shazeer}]{roberts2020much}
Adam Roberts, Colin Raffel, and Noam Shazeer. 2020.
\newblock How much knowledge can you pack into the parameters of a language
  model?
\newblock \emph{arXiv preprint arXiv:2002.08910}.

\bibitem[{Sang and De~Meulder(2003)}]{sang2003introduction}
Erik~F Sang and Fien De~Meulder. 2003.
\newblock Introduction to the conll-2003 shared task: Language-independent
  named entity recognition.
\newblock \emph{arXiv preprint cs/0306050}.

\bibitem[{Shuster et~al.(2019)Shuster, Ju, Roller, Dinan, Boureau, and
  Weston}]{shuster2019dialogue}
Kurt Shuster, Da~Ju, Stephen Roller, Emily Dinan, Y-Lan Boureau, and Jason
  Weston. 2019.
\newblock The dialogue dodecathlon: Open-domain knowledge and image grounded
  conversational agents.
\newblock \emph{arXiv preprint arXiv:1911.03768}.

\bibitem[{Surdeanu(2013)}]{surdeanu2013overview}
Mihai Surdeanu. 2013.
\newblock Overview of the tac2013 knowledge base population evaluation: English
  slot filling and temporal slot filling.
\newblock In \emph{TAC}.

\bibitem[{Surdeanu and Ji(2014)}]{surdeanu2014overview}
Mihai Surdeanu and Heng Ji. 2014.
\newblock Overview of the english slot filling track at the tac2014 knowledge
  base population evaluation.
\newblock In \emph{Proc. Text Analysis Conference (TAC2014)}.

\bibitem[{Talmor and Berant(2018)}]{talmor2018web}
Alon Talmor and Jonathan Berant. 2018.
\newblock \href {https://doi.org/10.18653/v1/N18-1059} {The web as a
  knowledge-base for answering complex questions}.
\newblock \emph{Proceedings of the 2018 Conference of the North {A}merican
  Chapter of the Association for Computational Linguistics: Human Language
  Technologies, Volume 1 (Long Papers)}, pages 641--651.

\bibitem[{Thorne and Vlachos(2020)}]{thorne2020avoiding}
James Thorne and Andreas Vlachos. 2020.
\newblock Avoiding catastrophic forgetting in mitigating model biases in
  sentence-pair classification with elastic weight consolidation.
\newblock \emph{arXiv preprint arXiv:2004.14366}.

\bibitem[{Thorne et~al.(2018{\natexlab{a}})Thorne, Vlachos, Christodoulopoulos,
  and Mittal}]{Thorne18Fever}
James Thorne, Andreas Vlachos, Christos Christodoulopoulos, and Arpit Mittal.
  2018{\natexlab{a}}.
\newblock {FEVER}: a large-scale dataset for fact extraction and verification.
\newblock In \emph{NAACL-HLT}.

\bibitem[{Thorne et~al.(2018{\natexlab{b}})Thorne, Vlachos, Cocarascu,
  Christodoulopoulos, and Mittal}]{thorne2018fact}
James Thorne, Andreas Vlachos, Oana Cocarascu, Christos Christodoulopoulos, and
  Arpit Mittal. 2018{\natexlab{b}}.
\newblock The fact extraction and verification (fever) shared task.
\newblock \emph{EMNLP}.

\bibitem[{Thorne et~al.(2019)Thorne, Vlachos, Cocarascu, Christodoulopoulos,
  and Mittal}]{thorne2019fever2}
James Thorne, Andreas Vlachos, Oana Cocarascu, Christos Christodoulopoulos, and
  Arpit Mittal. 2019.
\newblock The fever2. 0 shared task.
\newblock In \emph{Proceedings of the Second Workshop on Fact Extraction and
  VERification (FEVER)}, pages 1--6.

\bibitem[{Wang et~al.(2019)Wang, Pruksachatkun, Nangia, Singh, Michael, Hill,
  Levy, and Bowman}]{wang2019superglue}
Alex Wang, Yada Pruksachatkun, Nikita Nangia, Amanpreet Singh, Julian Michael,
  Felix Hill, Omer Levy, and Samuel Bowman. 2019.
\newblock Superglue: A stickier benchmark for general-purpose language
  understanding systems.
\newblock In \emph{Advances in Neural Information Processing Systems}, pages
  3261--3275.

\bibitem[{Wang et~al.(2018)Wang, Singh, Michael, Hill, Levy, and
  Bowman}]{wang2018glue}
Alex Wang, Amanpreet Singh, Julian Michael, Felix Hill, Omer Levy, and Samuel~R
  Bowman. 2018.
\newblock Glue: A multi-task benchmark and analysis platform for natural
  language understanding.
\newblock \emph{arXiv preprint arXiv:1804.07461}.

\bibitem[{Wolf et~al.(2019)Wolf, Debut, Sanh, Chaumond, Delangue, Moi, Cistac,
  Rault, Louf, Funtowicz et~al.}]{wolf2019huggingface}
Thomas Wolf, L~Debut, V~Sanh, J~Chaumond, C~Delangue, A~Moi, P~Cistac, T~Rault,
  R~Louf, M~Funtowicz, et~al. 2019.
\newblock Huggingface's transformers: State-of-the-art natural language
  processing.
\newblock \emph{ArXiv, abs/1910.03771}.

\bibitem[{Wolfson et~al.(2020)Wolfson, Geva, Gupta, Gardner, Goldberg, Deutch,
  and Berant}]{wolfson2020break}
Tomer Wolfson, Mor Geva, Ankit Gupta, Matt Gardner, Yoav Goldberg, Daniel
  Deutch, and Jonathan Berant. 2020.
\newblock Break it down: A question understanding benchmark.
\newblock \emph{Transactions of the Association for Computational Linguistics},
  8:183--198.

\bibitem[{Wu et~al.(2019)Wu, Petroni, Josifoski, Riedel, and
  Zettlemoyer}]{wu2019zero}
Ledell Wu, Fabio Petroni, Martin Josifoski, Sebastian Riedel, and Luke
  Zettlemoyer. 2019.
\newblock Zero-shot entity linking with dense entity retrieval.
\newblock \emph{arXiv preprint arXiv:1911.03814}.

\bibitem[{Xiong et~al.(2019)Xiong, Du, Wang, and
  Stoyanov}]{xiong2019pretrained}
Wenhan Xiong, Jingfei Du, William~Yang Wang, and Veselin Stoyanov. 2019.
\newblock Pretrained encyclopedia: Weakly supervised knowledge-pretrained
  language model.
\newblock \emph{arXiv preprint arXiv:1912.09637}.

\bibitem[{Yadav et~al.(2019)Yadav, Jain, Agrawal, Chattopadhyay, Singh, Jain,
  Singh, Lee, and Batra}]{EvalAI}
Deshraj Yadav, Rishabh Jain, Harsh Agrawal, Prithvijit Chattopadhyay, Taranjeet
  Singh, Akash Jain, Shiv~Baran Singh, Stefan Lee, and Dhruv Batra. 2019.
\newblock Evalai: Towards better evaluation systems for ai agents.
\newblock \emph{arXiv preprint arXiv:1902.03570}.

\bibitem[{Yang et~al.(2018)Yang, Qi, Zhang, Bengio, Cohen, Salakhutdinov, and
  Manning}]{yang2018hotpotqa}
Zhilin Yang, Peng Qi, Saizheng Zhang, Yoshua Bengio, William Cohen, Ruslan
  Salakhutdinov, and Christopher~D. Manning. 2018.
\newblock \href {https://doi.org/10.18653/v1/D18-1259} {{H}otpot{QA}: A dataset
  for diverse, explainable multi-hop question answering}.
\newblock \emph{Proceedings of the 2018 Conference on Empirical Methods in
  Natural Language Processing}, pages 2369--2380.

\end{thebibliography}
\bibliographystyle{acl_natbib}

\clearpage
\appendix
\section{Appendix}

\paragraph{Wikipedia Representation} We represent the KILT knowledge source as a collection of JSON records, one per Wikipedia page. Each record is assigned:
\begin{enumerate*}[label=(\roman*)] 
\item a unique Wikipedia id; 
\item a unique Wikipedia title;
\item a text field containing a list of strings, one for each paragraph, bulleted list item, and section header (for which we preserve the hierarchical structure);
\item a list of anchors elements, one for each hyperlink in the original page text, with span reference in the text field and page linked;
\item a list of categories;
\item a url redirecting to the original html for the page, with timestamp of the last page revision before the considered snapshot.
\end{enumerate*}

\paragraph{Datasets Mapping Details}

In \textit{FEVER}, often multiple pieces of knowledge must be combined to produce an output. For example, 30\% of claims have more than one equally-valid provenance and 16\% require the combination of multiple evidence spans.
% When annotators can't find evidence in the Wikipedia snapshot to support or refute the claims, they label them as not enough info. 
%claim Each claim is classified as supported, refuted or not enough info. Supported and refuted claims are accompanied by a set of Wikipedia sentences forming the necessary evidence for the judgment.
The second iteration~\cite[\emph{FEVER2.0},][]{thorne2019fever2} introduces a collection of  adversarial instances. %JT removed thorne2019adversarial
For KILT, we merge the two versions of FEVER into a single resource and consider only supported refuted claims. 
We exclude all claims classified as not having enough information since these instances have no evidence to assess the claim and cannot be mapped to the KILT knowledge source. Therefore we cannot asses whether such label is still appropriated given our snapshot. Moreover, we design KILT as an in-KB resource where each instance can be answered and corroborated by information in the knowledge source.

In the \textit{Zero Shot RE} dataset a set crowd-sourced template questions are defined for each relation --- for example, \emph{What is Albert Einstein’s alma mater?}.
%for the example above.
Each datapoint reports a Wikipedia sentence expressing the fact that we take as \emph{provenance}.
Some examples in the dataset are negative, obtained by matching a valid question and a random sentence, that likely does not contain the answer. 
To consider an open-domain version of this dataset and align the input/output with the KILT interface we reformatted this dataset, as follows:
\begin{enumerate*}[label=(\roman*)] 
\item exclude neagative pairs - since we consider the whole knowledge source (as opposite to a single sentence) as text all questions can be answered;
\item group template questions by the subject-relation pair, and create a single datapoint for each (\textit{input} as above);
\item randomly split the set of relations, in line with the original dataset, into three disjoint sets train (with 84 relations), dev (12 relations) and test (24 relations)---systems are tested on relations never seen during training;
\item use the subject entity as the query against Wikipedia titles for the first step of the mapping strategy, and
\item include all template questions in a \textit{meta} field.
\end{enumerate*}

For \textit{T-REx}, We filter out facts with more than 20 provenances, relations with less than 1000 facts, and merge all the facts for the same subject-relation pair (i.e., for 1-N and M-N relations there could be multiple valid answers), resulting in 113 relations and 2.3M facts. We include object aliases as equally valid answers and report in a \textit{meta} field subject aliases as well as all surface mentions for the subject, relation and object.
We randomly select 5k facts for both dev and test set.

To define an open-version of the \textit{Natural Questions} dataset we follow~\citet{lee2019latent} and (1) keep only questions with short answers and (2) discard all answers with more than five tokens.

To find answers in \textit{TriviaQA}, the original work used distant supervision: (1) find Wikipedia entities in the question with the TAGME entity linked~\cite{ferragina2011fast}; (2) search for the answer (and all Wikipedia aliases) in the corresponding page; (3) if the answer is found, add the page in the evidence documents.
Therefore, the documents are not guaranteed to contain evidence for the question-answer pair 
(but the authors estimate that they do 79.7\% of the time).

In \textit{ELI5} Evidence documents are automatically gathered, and we focus on the case where evidence documents are extracted from Wikipedia. However, as the original work first collected question-answer pairs from the subreddit \textit{Explain Like I'm Five}, the documents are not guaranteed to contain evidence. 

For \textit{Wizard of Wikipedia} we discard cases where the dataset does not contain provenance. Moreover, we consider a full open-domain setting where no topic is provided for the conversation and the model must search over all of Wikipedia for knowledge at each dialogue turn (rather than the provided knowledge candidates for each turn in the original dataset). We use the unseen split for dev and test.

\begin{figure}[t!]
    \centering
    \includegraphics[width=\linewidth]{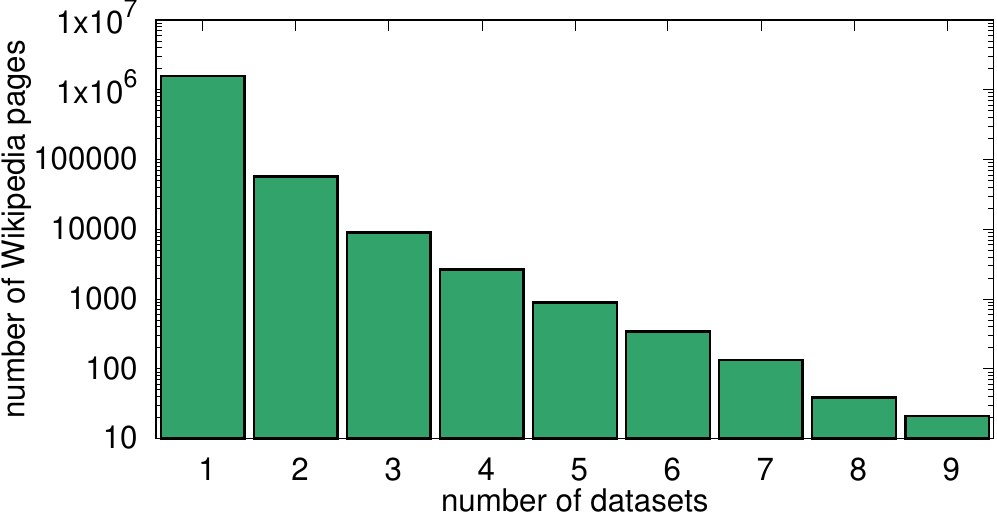}
    \caption{Number of pages vs number of dataset with knowledge in a page. 
    1,642,311 pages contains knowledge needed for KILT (\textasciitilde28\% of the knowledge source). 
    }
    \label{fig:number-pages}
\end{figure}

\paragraph{Performance Impact Of The Mapping Strategy}
We want to assess if the performance we obtain after mapping each dataset to a unified Wikipedia snapshot are in line with what reported in previous work. 
\citet{thorne2020avoiding} report a 2-way accuracy of 79.09 for the FEVER dev set when considering purely claims in input to a RoBERTa-based classifier~\cite{liu2019roberta}. Our dev set includes also the adversarial examples of FEVER 2.0, nevertheless the performance of BART are in line (80.67 dev, 78.93 test).
\citet{karpukhin2020dense} report 41.5 for EM on the open domain version of the NQ dev set\footnote{Reported as test results in \cite{karpukhin2020dense}}. With our setting, DPR achieves an on-par performance on the dev set, with a 42.58 EM (50.43 F1-score). Results on our brand new NQ test set are 3/4 points lower for EM and F1-score than dev results. 
%
%\fabio{TriviaQA?}
%
%\fabio{say soimething about HotpotQA, that an excrtactive approach is not enough for provenance, hence 0.0 performance in KILT-score}
We don't evaluate multi-hop specific baselines on KILT but the current best F1-score for HotpotQA is 75.43 according to the official leadearboard\footnote{\url{https://hotpotqa.github.io}}, that is quite far from what achieved by our general solutions. 
%
%\fabio{zero shot?}
%
BLINK results are in line with what reported in the GitHub repository\footnote{\url{https://github.com/facebookresearch/BLINK}} for all three entity linking datasets.
The Tranformer MemNet of~\citet{dinan2018wizard} achieves a F1-score of 14.3 on the original version of the WW dataset while 11.5 in our setting, probably because in KILT we consider an harder open-domain setting.
%where conversation topic and candidate Wikipedia pages are not provided.

%\subsection{Baselines}

\paragraph{Retrieval Baselines}
%One important component we evaluate in KILT is the ability to produce accurate provenance information from Wikipedia (in the form of a page) for each prediction. 
The ability to retrieve relevant documents from Wikipedia given an input is an important aspect we assess in KILT. A system should select only the relevant knowledge needed for the task, without redundant or excess information.
A way to surface such knowledge is using a dedicated retrieval system.
We consider three off-the-shelf retrievers and investigate drastically different retrieval paradigms:
\begin{enumerate*}[label=(\roman*)] 
\item \emph{Tf-idf} with the DrQA Document Retriever~\cite{chen2017reading}---traditional page-level sparse vector space retrieval model;
\item \emph{DPR}~\cite{karpukhin2020dense}---a modern passage-level retrieval solution using dense representations;
\item A combination of \emph{BLINK}~\cite{wu2019zero} and \emph{flair}~\cite{akbik2019flair}---retrieval solution that ranks pages according to  entities in the input.
\end{enumerate*}
 %Additional details are in the appendix.

%\paragraph{Retrieval baselines additional details.}
The DrQA Document Retriever combines bigram hashing and TF-IDF matching to return relevant Wikipedia pages given an input.
DPR splits each Wikipedia page into disjoint 100-word passages\footnote{22,220,793 passages in the KILT knowledge source. Following~\citet{karpukhin2020dense} we don't consider Wikipedia bulleted lists in the text.} and encodes passages and inputs with a BERT-based bi-encoder to perform dense Maximum Inner Product Search. 
The BLINK entity linking system uses a BERT-based bi-encoder to encode each Wikipedia page as well as each input, where a single entity mention is tagged. Final results are refined with a BERT-based cross-encoder. To use BLINK for retrieval, we look for entity mentions in each input with flair, then use BLINK to return a ranked list of Wikipedia pages for each entity mention. When multiple entities are identified in the input, we merge results and sort by score. The input string might not contain tags.
For all systems, we use the index created on the KILT knowledge source.
%\paragraph{Multi-task DPR.}

We also experiment with multi-tasking, by jointly training a single DPR model on all KILT training data. We use uniform sampling to balance the datasets.
In particular, the Multi-task variant of DPR is a single dense passage retriever, trained jointly on the union of TQA, NQ, HoPo, FEV, zsRE, AY2, T-REx and WoW. In order to avoid large datasets, such as T-REx, from having an oversize effect, we resample all datasets uniformly, such that every training epoch contains 150k samples from each task. Batches are formed from a single dataset at a time, iterating through the various datasets in a round-robin fashion.

\paragraph{Task-specific Baselines}
\label{sec:baselie-task}

Approaches to the KILT Benchmark should be able to generalize to many different tasks, as developing model architectures that can represent knowledge generally is a valuable direction. However, several tasks may benefit from dedicated architectures designed for them. 
%

%\paragraph{Fact Checking}
For \emph{fact checking}, we consider \emph{NSMN}~\cite{nie2019combining}, the highest scoring system from the FEVER shared task~\cite{thorne2018fact}. %https://www.aclweb.org/anthology/W18-5501/
%\jt{(TODO: aside) also the highest scoring pre-trained system on KILT with an open-source implementation}. 
%It is a pipeline of three neural semantic matching models (based on the Enhanced LSTM architecture~\cite{chen-etal-2017-enhanced}) for document retrieval, sentence selection (binary classification for passage relevance), and claim verification. 
We use the public model\footnote{available at \url{https://github.com/easonnie/combine-FEVER-NSMN}} pre-trained on FEVER, and consider not enough information predictions as false.
%\paragraph{BERT+DPR FEVER baseline}
Moreover, we develop a fact checking baseline that combines a BERT-base classifier with passages returned from DPR where the claim and retrieved passage are input.
%We evaluate this model on fact checking, question answering, and slot filling tasks. 
%The model input is a \textit{<claim,passage>} pair using passages returned from DPR, and the output i
%We train a 3-way classifier where the retrieved passage and claim are input.
The classifier is trained to label the claim-passage pair as supported or refuted with an additional neutral class for negative-sampled unrelated passages. 
Unrelated passages are sampled from two sources: (1) DPR-retrieved passages from pages that are not in the list of pages in the instance's provenance and (2) passages sampled uniformly at random from pages in the instance's provenance. 
At inference, we classify the first sentence of the Wikipedia pages retrieved by the top-100 DPR passages against the claim. 
Using pages labelled as supported or refuted, we label the claim through majority voting.
For claim provenance, we re-rank passages by probability according to this label. 

For \emph{Open Domain QA} and {Slot Filling}, we use DPR combined with the pre-trained BERT-based extractive reading comprehension model of~\citet{karpukhin2020dense}. We use the model pretrained on TriviaQA for HotpotQA and the model pre-trained on Natural Questions for Zero Shot RE.
We reduce the slot filling problem to question answering, by using the specified template questions. We consider a single random template question per subject-relation during inference.

For \emph{Entity Linking}, we consider \emph{BLINK}.

For \emph{Dialogue}, we consider the Generative \emph{Transformer MemNet}~\cite{dinan2018wizard} that encodes the dialogue history and knowledge to generates the next utterance. We use the pre-trained version available in ParlAI~\cite{miller2017parlai}.
Finally, to test the performance of combining BART and DPR on FEVER, we develop a classifier that uses these---full description in the appendix.

\paragraph{General Baselines}
A main motivation of the KILT Benchmark is to enable a unified approach towards a wide range of knowledge-intensive tasks. We analyze existing general architectures that can be used as a baseline for multiple tasks in KILT.

%\paragraph{Implicit Knowledge Access}
Large pre-trained sequence-to-sequence models such as BART~\cite{Lewis2019BARTDS} and T5~\cite{2019t5} implicitly store a surprising amount of knowledge in their parameters~\cite{petroni2019language}. We treat all KILT tasks as generative, relying on the knowledge accumulated by the model while pre-training, with no retrieval (similarly to \citet{roberts2020much}). We finetune pre-trained variants on all KILT tasks, using \texttt{fairseq}~\cite{ott2019fairseq} for BART and Huggingface’s Transformer~\cite{wolf2019huggingface} for T5.
%and report end-to-end F1 scores (see Table~\ref{tab:results}). 
%As answers are generated solely from the input with no explicit retrieval, there is no straightforward way to access provenance for each prediction. We therefore do not report KILT scores for these.
%We do not report provenance metrics for these baselines.

%\paragraph{BART.} BART is a Transformer-based \cite{vaswani2017attention} denoising autoencoder which first corrupts the original text using a set of noise functions and then learns to recover the input. We fine-tune the BART-large model for each task with the  \texttt{fairseq} library \cite{ott2019fairseq}. 
%\paragraph{T5} \fabio{T5}

%\paragraph{Explicit knowledge access}
%As detailed in Section~\ref{sec:baselie-task} \piktus{Verify this once that section is actually written}, combining a retriever, explicitly surfacing external knowledge, with a reader has served as a successful strategy for a variety of knowledge-intensive tasks. However, it has mostly been applied to solve isolated problems, mainly in the domain of question-answering.
A natural way to boost performance is to incorporate an explicit knowledge mechanism.
%\paragraph{BART+DPR}  
For our BART+DPR baseline,
%, we build a custom DPR index~\cite{karpukhin2020dense} on top of the KILT knowledge source. Before training, w
we follow \citet{petroni2020context} to retrieve and prepend the top-3 passages from DPR for each input sample and use context-enhanced training data to fine-tune a BART model.
We use the DPR rank when reporting provenance for all except entity linking tasks. For entity linking, we report the Wikipedia id of the page whose title exactly matches the predicted string. 

Recently, state-of-the-art results on a wide range of NLP tasks have been achieved by combining a trainable retrieval step with language modeling or generation~\cite{guu2020realm,lewis2020pre}. We experiment with fine-tuning RAG~\cite{lewis2020retrievalaugmented} on KILT tasks, establishing a strong baseline on all of them.
%\paragraph{RAG.} 
RAG combines a DPR retriever with a BART generator, however, unlike in the case of our previous baseline, RAG back-propagates to the retriever's input encoder, learning to adapt the input embedding to retrieve more relevant results. 
%We intitialize RAG training with the same KILT-specific DPR instance we built for the BART+DPR baseline and fine-tune it on KILT tasks. 
At every generation step we retrieve top-5 passages and use them as provenance.

\begin{table*}[t!]
    \centering
\resizebox{\textwidth}{!}{    
    \fontsize{8.4}{10.1}\selectfont \setlength{\tabcolsep}{0.5em}
    \begin{tabular}{rccccccccc}
        \toprule
        \makecell{Dataset\\Label} & Multi-hop & \makecell{Average\\Provenance\\Size (APS)} & \makecell{Average\\Provenance\\Number (APN)} & \makecell{Average\\Provenance\\Pages (APP)}  &
        \makecell{Average\\Answers\\Number (AAN)}  & \makecell{Train\\Size} & \makecell{Dev\\Size} & \makecell{Test\\Size} \\
        \midrule
        \textbf{FEV}  & x & 1.12 & 1.35 & 1.13 & 1  & 104,966 & 10,444 & 10,100 \\
        \textbf{AY2}  & & 1 & 1 & 1 & 1  & 18,395 & 4,784 & 4,463  \\
        \textbf{WnWi} & & 1 & 1 & 1 & 1 & - & 3,396 & 3,376  \\
        \textbf{WnCw} & & 1 & 1 & 1 & 1 & - & 5,599 & 5,543  \\
        \textbf{T-REx}  & & 1 & 1.68 & 1.26 & 5.29 & 2,284,168 & 5,000 & 5,000  \\
        \textbf{zsRE}   & & 1 & 1 & 1 & 1 & 147,909 & 3,724 & 4,966  \\
        \textbf{NQ}  & & 1 & 3.22 & 1.57  & 2.08  & 87,372 & 2,837 & 1,444\\
        \textbf{HoPo}  & x & 2.4 & 1 & 2 & 1 & 88,869 & 5,600 & 5,569 \\
        \textbf{TQA}  & & 1 & 3.39 & 1.68 & 28.67  & 61,844 & 5,359 & 6,586  \\
        \textbf{ELI5}  & & 1 & 1.21 & 1.18 & 4.69  & 272,634 & 1,507 & 600 \\
        \textbf{WoW}  & & 1 & 1 & 1 & 1 & 63,734 & 3,054 & 2,944  \\
        \midrule
        \multicolumn{6}{c}{\textit{Total}} & 3,129,891 & 51,460 & 50,736 & \\
        \bottomrule
    \end{tabular}
}
    \caption{Datasets statistics. APS refers to the average number of textual spans in each provenance set---for most of the datasets a single span is sufficient to provide enough evidence while FEV and HoPo might require more (hence they require multi-hop reasoning). APN indicates the average number of equally valid provenance sets for each instance while APP the average number of Wikipedia pages overall in the provenance (note that multiple spans might refer to the same Wikipedia page). Finally AAN reports the average number of equally valid gold answers per instance. We additionally report the size of the train, dev and test split for each dataset.}
    \label{tab:datasetsstats}
\end{table*}

\begin{figure*}[!ht]
\centering
\begin{lstlisting}[language=Python,firstnumber=1]
{'id': # original data point id if available otherwise unique id
 'input': # question / claim / sentence / etc
 'output': [ # each element might contain an answer, a provenance or both
    {
    'answer': # answer in textual form
    'provenance': [
        # evidence set for the answer from the KILT knowledge source
        {
            'wikipedia_id':  # *mandatory* 
            'title': 
            'section': 
            'start_paragraph_id': 
            'start_character': 
            'end_paragraph_id':
            'end_character': 
            'bleu_score': # wrt original evidence
            'meta': # dataset/task specific
        }
        ] 
      }
    ]
 'meta': # dataset/task specific
 }
\end{lstlisting}
    \caption{KILT datasets' interface. Each dataset is represented as a JSON Line file. The Figure shows the pseudo-JSON structure for each record in the files.}
    \label{fig:interface}
\end{figure*}

\begin{table*}[ht]
\centering
\begin{tabular}{l@{\hskip 2em}rrrr}
 \toprule
model & \textbf{R-Precision} & \textbf{Recall@5} & \textbf{Accuracy}  & \textbf{KILT-AC}  \\
\midrule
\multicolumn{5}{c}{test} \\
\midrule
BART & 0.0 & 0.0 & 78.93 & 0.0 \\ 
T5 & 0.0 & 0.0 & 76.3 & 0.0 \\ 
NSMN & 49.24 & 70.16 & 66.1 & 41.88 \\ 
BART + DPR & 55.33 & 74.29 & 86.74 & 47.68 \\ 
RAG & 61.94 & 75.55 & 86.31 & 53.45 \\ 
BERT + DPR & 72.93 & 73.52 & 69.68 & 58.58 \\ 
\midrule
\multicolumn{5}{c}{dev} \\
\midrule
BART & 0.0 & 0.0 & 80.67 & 0.0 \\ 
BART + DPR & 55.46 & 73.84 & 88.11 & 48.25 \\ 
RAG & 63.5 & 76.1 & 87.7 & 55.47 \\ 
\bottomrule
\end{tabular}
\caption{FEVER}
\label{tab:FEV}
\end{table*}

\begin{table*}[ht]
\centering
\begin{tabular}{l@{\hskip 2em}rrrr}
 \toprule
model & \textbf{R-Precision} & \textbf{Recall@5} & \textbf{Accuracy}  & \textbf{KILT-AC}  \\
\midrule
\multicolumn{5}{c}{test} \\
\midrule
RAG & 72.62 & 72.62 & 72.62 & 72.62 \\ 
T5 & 74.05 & 74.05 & 74.05 & 74.05 \\ 
BART + DPR & 75.49 & 75.49 & 75.49 & 75.49 \\ 
BART & 77.55 & 77.55 & 77.55 & 77.55 \\ 
BLINK & 81.54 & 94.73 & 81.54 & 81.54 \\ 
\midrule
\multicolumn{5}{c}{dev} \\
\midrule
RAG & 77.4 & 77.47 & 77.4 & 77.4 \\ 
T5 & 81.84 & 81.84 & 81.84 & 81.84 \\ 
BART & 86.62 & 86.62 & 86.62 & 86.62 \\ 
\bottomrule
\end{tabular}
\caption{AIDA CoNLL-YAGO}
\label{tab:AY2}
\end{table*}

\begin{table*}[ht]
\centering
\begin{tabular}{l@{\hskip 2em}rrrr}
 \toprule
model & \textbf{R-Precision} & \textbf{Recall@5} & \textbf{Accuracy}  & \textbf{KILT-AC}  \\
\midrule
\multicolumn{5}{c}{test} \\
\midrule
BART + DPR & 45.2 & 45.2 & 45.2 & 45.2 \\ 
BART & 45.91 & 45.91 & 45.91 & 45.91 \\ 
T5 & 47.13 & 47.13 & 47.13 & 47.13 \\ 
RAG & 48.07 & 48.07 & 48.07 & 48.07 \\ 
BLINK & 80.24 & 91.47 & 80.24 & 80.24 \\ 
\midrule
\multicolumn{5}{c}{dev} \\
\midrule
BART + DPR & 44.96 & 44.96 & 44.96 & 44.96 \\ 
T5 & 47.35 & 47.35 & 47.35 & 47.35 \\ 
BART & 47.91 & 47.91 & 47.91 & 47.91 \\ 
RAG & 49.0 & 49.0 & 49.0 & 49.0 \\ 
\bottomrule
\end{tabular}
\caption{WNED-WIKI}
\label{tab:WnWi}
\end{table*}

\begin{table*}[ht]
\centering
\begin{tabular}{l@{\hskip 2em}rrrr}
 \toprule
model & \textbf{R-Precision} & \textbf{Recall@5} & \textbf{Accuracy}  & \textbf{KILT-AC}  \\
\midrule
\multicolumn{5}{c}{test} \\
\midrule
BART + DPR & 46.87 & 46.87 & 46.87 & 46.87 \\
RAG & 47.61 & 47.61 & 47.61 & 47.61 \\
BART & 49.16 & 49.16 & 49.16 & 49.16 \\ 
T5 & 49.29 & 49.29 & 49.29 & 49.29 \\
BLINK & 68.77 & 81.78 & 68.77 & 68.77 \\ 
\midrule
\multicolumn{5}{c}{dev} \\
\midrule
BART + DPR & 45.7 & 45.7 & 45.7 & 45.7 \\
T5 & 46.58 & 46.58 & 46.58 & 46.58 \\ 
RAG & 46.7 & 46.7 & 46.7 & 46.7 \\ 
BART & 48.01 & 48.01 & 48.01 & 48.01 \\ 
\bottomrule
\end{tabular}
\caption{WNED-CWEB}
\label{tab:WnCw}
\end{table*}

\begin{table*}[ht]
\centering
\begin{tabular}{l@{\hskip 2em}rrrrrr}
 \toprule
model & \textbf{R-Precision} & \textbf{Recall@5} & \textbf{Accuracy} & \textbf{F1}  & \textbf{KILT-AC}  & \textbf{KILT-F1}  \\
\midrule
\multicolumn{7}{c}{test} \\
\midrule
BART & 0.0 & 0.0 & 45.06 & 49.24 & 0.0 & 0.0 \\ 
T5 & 0.0 & 0.0 & 43.56 & 50.61 & 0.0 & 0.0 \\ 
BART + DPR & 13.26 & 17.04 & 59.16 & 62.76 & 11.12 & 11.41 \\ 
RAG & 28.68 & 33.04 & 59.2 & 62.96 & 23.12 & 23.94 \\ 
\midrule
\multicolumn{7}{c}{dev} \\
\midrule
BART & 0.0 & 0.0 & 43.84 & 48.25 & 0.0 & 0.0 \\ 
T5 & 0.0 & 0.0 & 47.24 & 51.73 & 0.0 & 0.0 \\ 
BART + DPR & 13.62 & 16.93 & 56.7 & 60.19 & 11.56 & 11.87 \\ 
RAG & 29.26 & 33.69 & 61.48 & 65.03 & 25.4 & 26.22 \\ 
\bottomrule
\end{tabular}
\caption{T-REx}
\label{tab:TREx}
\end{table*}

\begin{table*}[ht]
\centering
\begin{tabular}{l@{\hskip 2em}rrrrrr}
 \toprule
model & \textbf{R-Precision} & \textbf{Recall@5} & \textbf{Accuracy} & \textbf{F1}  & \textbf{KILT-AC}  & \textbf{KILT-F1}  \\
\midrule
\multicolumn{7}{c}{test} \\
\midrule
BART & 0.0 & 0.0 & 9.14 & 12.21 & 0.0 & 0.0 \\ 
T5 & 0.0 & 0.0 & 9.02 & 13.52 & 0.0 & 0.0 \\ 
BERT + DPR & 40.11 & 40.11 & 6.93 & 37.28 & 4.47 & 27.09 \\ 
BART + DPR & 28.9 & 39.21 & 30.43 & 34.47 & 18.91 & 20.32 \\ 
RAG & 53.73 & 59.52 & 44.74 & 49.95 & 36.83 & 39.91 \\ 
\midrule
\multicolumn{7}{c}{dev} \\
\midrule
BART & 0.0 & 0.0 & 3.03 & 12.61 & 0.0 & 0.0 \\ 
T5 & 0.0 & 0.0 & 1.58 & 10.8 & 0.0 & 0.0 \\ 
BART + DPR & 45.6 & 58.49 & 34.96 & 44.79 & 29.08 & 32.85 \\ 
RAG & 65.36 & 73.07 & 47.42 & 57.98 & 42.64 & 48.35 \\ 
\bottomrule
\end{tabular}
\caption{Zero Shot RE}
\label{tab:zsRE}
\end{table*}

\begin{table*}[ht]
\centering
\begin{tabular}{l@{\hskip 2em}rrrrrr}
 \toprule
model & \textbf{R-Precision} & \textbf{Recall@5} & \textbf{EM} & \textbf{F1}  & \textbf{KILT-EM}  & \textbf{KILT-F1}  \\
\midrule
\multicolumn{7}{c}{test} \\
\midrule
BART & 0.0 & 0.0 & 21.75 & 28.69 & 0.0 & 0.0 \\ 
T5 & 0.0 & 0.0 & 19.6 & 27.73 & 0.0 & 0.0 \\ 
BART + DPR & 54.29 & 65.52 & 41.27 & 49.54 & 30.06 & 34.72 \\ 
BERT + DPR & 60.66 & 46.79 & 38.64 & 47.09 & 31.99 & 37.58 \\ 
RAG & 59.49 & 67.06 & 44.39 & 52.35 & 32.69 & 37.91 \\  
\midrule
\multicolumn{7}{c}{dev} \\
\midrule
BART & 0.0 & 0.0 & 26.15 & 32.06 & 0.0 & 0.0 \\ 
T5 & 0.0 & 0.0 & 25.2 & 31.88 & 0.0 & 0.0 \\ 
BART + DPR & 54.25 & 64.99 & 45.05 & 52.98 & 31.62 & 35.84 \\ 
BERT + DPR & 60.03 & 45.06 & 42.58 & 50.43 & 35.32 & 39.84 \\ 
RAG & 60.31 & 65.47 & 48.78 & 56.1 & 36.31 & 40.64 \\  
\bottomrule
\end{tabular}
\caption{Natural Questions}
\label{tab:NQ}
\end{table*}

\begin{table*}[ht]
\centering
\begin{tabular}{l@{\hskip 2em}rrrrrr}
 \toprule
model & \textbf{R-Precision} & \textbf{Recall@5} & \textbf{EM} & \textbf{F1}  & \textbf{KILT-EM}  & \textbf{KILT-F1}  \\
\midrule
\multicolumn{7}{c}{test} \\
\midrule
BART & 0.0 & 0.0 & 15.37 & 21.97 & 0.0 & 0.0 \\ 
T5 & 0.0 & 0.0 & 12.64 & 19.57 & 0.0 & 0.0 \\  
BERT + DPR & 25.04 & 10.4 & 11.29 & 17.35 & 0.74 & 1.26 \\ 
BART + DPR & 25.04 & 10.4 & 25.18 & 34.07 & 1.96 & 2.53 \\ 
RAG & 30.59 & 12.59 & 26.97 & 36.03 & 3.21 & 4.1 \\ 
\midrule
\multicolumn{7}{c}{dev} \\
\midrule
BART & 0.0 & 0.0 & 16.86 & 23.81 & 0.0 & 0.0 \\ 
T5 & 0.0 & 0.0 & 12.66 & 19.74 & 0.0 & 0.0 \\ 
BERT + DPR & 24.62 & 10.7 & 10.82 & 16.96 & 0.96 & 1.34 \\ 
BART + DPR & 24.62 & 10.7 & 25.75 & 35.2 & 1.96 & 2.46 \\ 
RAG & 30.76 & 12.29 & 27.68 & 37.37 & 3.14 & 3.87 \\  
\bottomrule
\end{tabular}
\caption{HotpotQA}
\label{tab:HotpotQA}
\end{table*}

\begin{table*}[ht]
\centering
\begin{tabular}{l@{\hskip 2em}rrrrrr}
 \toprule
model & \textbf{R-Precision} & \textbf{Recall@5} & \textbf{EM} & \textbf{F1}  & \textbf{KILT-EM}  & \textbf{KILT-F1}  \\
\midrule
\multicolumn{7}{c}{test} \\
\midrule
BART & 0.0 & 0.0 & 32.39 & 39.85 & 0.0 & 0.0 \\ 
T5 & 0.0 & 0.0 & 18.11 & 27.83 & 0.0 & 0.0 \\ 
BART + DPR & 44.49 & 56.99 & 58.55 & 67.79 & 31.4 & 35.34 \\ 
BERT + DPR & 43.4 & 31.45 & 70.38 & 74.41 & 34.48 & 36.28 \\ 
RAG & 48.68 & 57.13 & 71.27 & 75.88 & 38.13 & 40.15 \\  
\midrule
\multicolumn{7}{c}{dev} \\
\midrule
BART & 0.0 & 0.0 & 32.54 & 39.58 & 0.0 & 0.0 \\ 
T5 & 0.0 & 0.0 & 25.79 & 33.72 & 0.0 & 0.0 \\ 
BERT + DPR & 40.87 & 29.96 & 70.24 & 74.21 & 32.9 & 34.48 \\ 
BART + DPR & 45.36 & 56.72 & 59.28 & 68.31 & 32.56 & 36.36 \\ 
RAG & 49.26 & 56.93 & 61.73 & 67.12 & 36.13 & 38.71 \\  
\bottomrule
\end{tabular}
\caption{TriviaQA}
\label{tab:TQA}
\end{table*}

\begin{table*}[ht]
\centering
\begin{tabular}{l@{\hskip 2em}rrrrrr}
 \toprule
model & \textbf{R-Precision} & \textbf{Recall@5} & \textbf{Rouge-L} & \textbf{F1}  & \textbf{KILT-RL}  & \textbf{KILT-F1}  \\
\midrule
\multicolumn{7}{c}{test} \\
\midrule
T5 & 0.0 & 0.0 & 19.08 & 16.1 & 0.0 & 0.0 \\ 
BART & 0.0 & 0.0 & 20.55 & 19.23 & 0.0 & 0.0 \\ 
RAG & 11.0 & 22.92 & 14.05 & 14.51 & 1.69 & 1.79 \\ 
BART + DPR & 10.67 & 26.92 & 17.41 & 17.88 & 1.9 & 2.01 \\ 
\midrule
\multicolumn{7}{c}{dev} \\
\midrule
T5 & 0.0 & 0.0 & 21.02 & 18.36 & 0.0 & 0.0 \\ 
BART & 0.0 & 0.0 & 22.69 & 22.19 & 0.0 & 0.0 \\ 
RAG & 16.39 & 27.27 & 16.11 & 17.24 & 2.65 & 2.88 \\ 
BART + DPR & 16.32 & 21.11 & 18.53 & 18.75 & 2.87 & 2.89 \\ 
\bottomrule
\end{tabular}
\caption{ELI5}
\label{tab:ELI5}
\end{table*}

\begin{table*}[ht]
\centering
\begin{tabular}{l@{\hskip 2em}rrrrrr}
 \toprule
model & \textbf{R-Precision} & \textbf{Recall@5} & \textbf{Rouge-L} & \textbf{F1}  & \textbf{KILT-RL}  & \textbf{KILT-F1}  \\
\midrule
\multicolumn{7}{c}{test} \\
\midrule
BART & 0.0 & 0.0 & 11.77 & 12.86 & 0.0 & 0.0 \\ 
T5 & 0.0 & 0.0 & 12.40 & 13.53 & 0.0 & 0.0 \\ 
TransMemNet & 18.35 & 18.35 & 10.11 & 11.85 & 1.85 & 2.2 \\
BART + DPR & 25.46 & 55.1 & 13.23 & 15.19 & 3.71 & 4.37 \\ 
RAG & 57.75 & 74.61 & 11.57 & 13.11 & 7.59 & 8.75 \\   
\midrule
\multicolumn{7}{c}{dev} \\
\midrule
BART & 0.0 & 0.0 & 12.25 & 13.77 & 0.0 & 0.0 \\ 
T5 & 0.0 & 0.0 & 12.36 & 13.15 & 0.0 & 0.0 \\ 
BART + DPR & 0.0 & 0.0 & 13.48 & 15.51 & 0.0 & 0.0 \\ 
RAG & 46.66 & 66.57 & 11.76 & 13.28 & 6.72 & 7.5 \\  
\bottomrule
\end{tabular}
\caption{Wizard of Wikipedia}
\label{tab:WoW}
\end{table*}

\begin{figure*}[!ht]
\centering
\begin{lstlisting}[language=Python,firstnumber=1]
input: 'SOCCER - UNCAPPED PLAYERS CALLED TO FACE MACEDONIA . '[SE0]'BUCHAREST'[EE0]' 1996-12-06  '[SE1]'Romania'[EE1]' trainer '[SE2]'Anghel Iordanescu'[EE2]' called up three uncapped players on Friday in his squad to face '[SE3]'Macedonia'[EE3]' next week in a '[SE4]'World Cup'[EE4]' qualifier .  Midfielder Valentin Stefan and striker '[SE5]'Viorel Ion'[EE5]' of Otelul Galati and defender '[SE6]'Liviu Ciobotariu'[EE6]' of National Bucharest are the newcomers for the '[SE7]'European'[EE7]' group eight clash in '[SE8]'Macedonia'[EE8]' on December 14 .  Iordanescu said he had picked them because of their good performances in the domestic championship in which National Bucharest are top and Otelul Galati third . "  I think it s fair to give them a chance , " he told reporters .  League title-holders Steaua Bucharest , who finished bottom of their Champions  League group in the '[SE9]'European Cup'[EE9]', have only two players in the squad .  Attacking midfielder '[SE10]'Adrian Ilie'[EE10]', who recently moved from Steaua to Turkish club '[SE11]'Galatasaray'[EE11]', is ruled out after two yellow-card offences .  Squad :  Goalkeepers - '[SE12]'Bogdan Stelea'[EE12]', '[SE13]'Florin Prunea'[EE13]'.  Defenders - '[SE14]'Dan Petrescu'[EE14]', '[SE15]'Daniel Prodan'[EE15]', Anton Dobos , Cornel Papura , '[SE16]'Liviu Ciobotariu'[EE16]', Tibor Selymess , '[SE17]'Iulian Filipescu'[EE17]'.  Midfielders - '[SE18]'Gheorghe Hagi'[EE18]', '[SE19]'Gheorghe Popescu'[EE19]', '[SE20]'Constantin Galca'[EE20]' , Valentin Stefan , '[SE21]'Basarab Panduru'[EE21]', '[SE22]'Dorinel Munteanu'[EE22]', Ovidiu Stinga .  Forwards - Ioan Vladoiu , '[SE23]'Gheorghe Craioveanu'[EE23]', '[SE24]'Ionel Danciulescu'[EE24]', '[SE25]'Viorel Ion'[EE25]' .  REUTER'

BART predictions:
v E0: 'Bucharest' -> https://en.wikipedia.org/wiki/Bucharest
x E1: 'Romania' (gold:'Romania national football team')
x E2: 'Anghel Iordanescu' (gold:'Anghel Iordanescu')
v E3: 'North Macedonia national football team' -> https://en.wikipedia.org/wiki/North_Macedonia_national_football_team
x E4: '1998 FIFA World Cup' (gold:'FIFA World Cup')
v E5: 'Viorel Ion' -> https://en.wikipedia.org/wiki/Viorel_Ion
v E6: 'Liviu Ciobotariu' -> https://en.wikipedia.org/wiki/Liviu_Ciobotariu
v E7: 'Europe' -> https://en.wikipedia.org/wiki/Europe
v E8: 'North Macedonia' -> https://en.wikipedia.org/wiki/North_Macedonia
v E9: 'UEFA Champions League' -> https://en.wikipedia.org/wiki/UEFA_Champions_League
v E10: 'Adrian Ilie' -> https://en.wikipedia.org/wiki/Adrian_Ilie
v E11: 'Galatasaray S.K. (football)' -> https://en.wikipedia.org/wiki/Galatasaray_S.K._(football)
v E12: 'Bogdan Stelea' -> https://en.wikipedia.org/wiki/Bogdan_Stelea
v E13: 'Florin Prunea' -> https://en.wikipedia.org/wiki/Florin_Prunea
v E14: 'Dan Petrescu' -> https://en.wikipedia.org/wiki/Dan_Petrescu
v E15: 'Daniel Prodan' -> https://en.wikipedia.org/wiki/Daniel_Prodan
v E16: 'Liviu Ciobotariu' -> https://en.wikipedia.org/wiki/Liviu_Ciobotariu
v E17: 'Iulian Filipescu' -> https://en.wikipedia.org/wiki/Iulian_Filipescu
v E18: 'Gheorghe Hagi' -> https://en.wikipedia.org/wiki/Gheorghe_Hagi
v E19: 'Gheorghe Popescu' -> https://en.wikipedia.org/wiki/Gheorghe_Popescu
x E20: 'Constantinos Galca' (gold:'Constantin Galca')
v E21: 'Basarab Panduru' -> https://en.wikipedia.org/wiki/Basarab_Panduru
v E22: 'Dorinel Munteanu' -> https://en.wikipedia.org/wiki/Dorinel_Munteanu
v E23: 'Gheorghe Craioveanu' -> https://en.wikipedia.org/wiki/Gheorghe_Craioveanu
x E24: 'Ion Danciulescu' (gold:'Ionel Danciulescu')
v E25: 'Viorel Ion' -> https://en.wikipedia.org/wiki/Viorel_Ion

F1-score = 87.52
KILT-F1-score = 21/26 = 80.77
EM = 21/26 = 80.77
KILT-EM-score = 21/26 = 80.77

\end{lstlisting}
    \caption{Entity linking BART predictions, schematic of 25 input-output pairs condensed, in each one a single entity in tagged. }
    \label{fig:entitylinking}
\end{figure*}

\begin{figure*}[ht]

\begin{subfigure}{.47\textwidth}
  \centering
  % include first image
  \includegraphics[width=\linewidth]{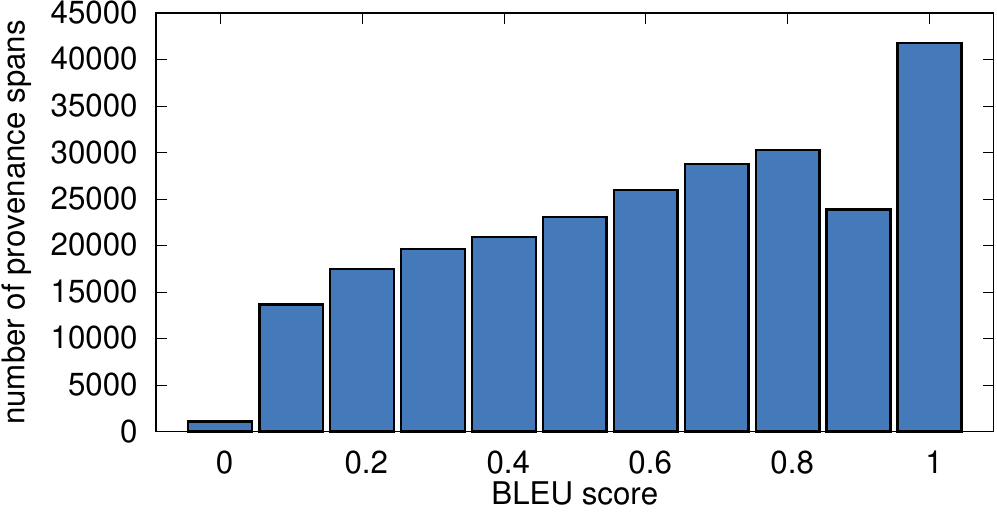}  
  \caption{FEVER, dev data discarded 26.03\% (3675), test data discarded 27.7\% (3869).}
  \label{fig:sub-FE}
\end{subfigure}\hspace{.05\textwidth}% Space between image B and C
\begin{subfigure}{.47\textwidth}
  \centering
  % include second image
  \includegraphics[width=\linewidth]{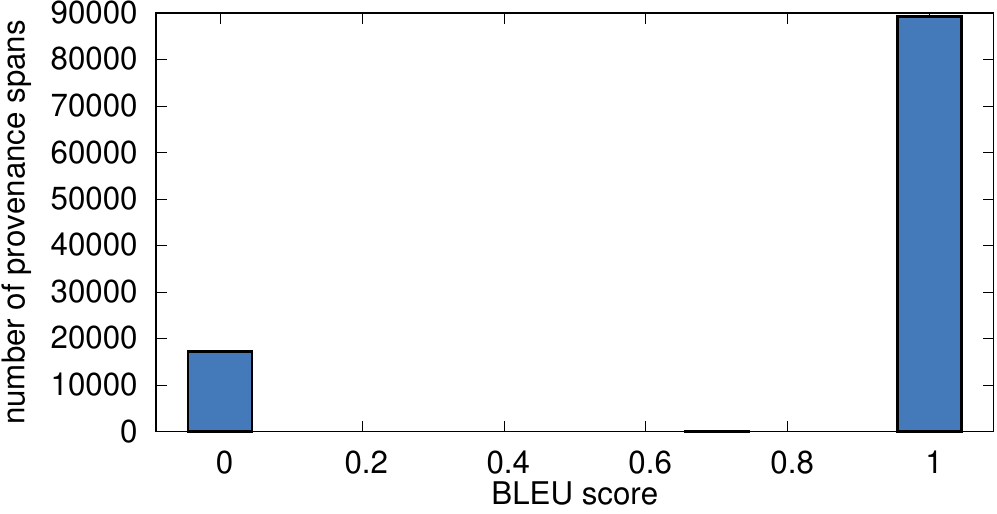}  
  \caption{Natural Questions, dev data discarded 16.12\% (595), test data discarded 15.59\% (287).}
  \label{fig:sub-NQ}
\end{subfigure}

\vspace{2em}% Space between image B and C

\begin{subfigure}{.47\textwidth}
  \centering
  % include first image
  \includegraphics[width=\linewidth]{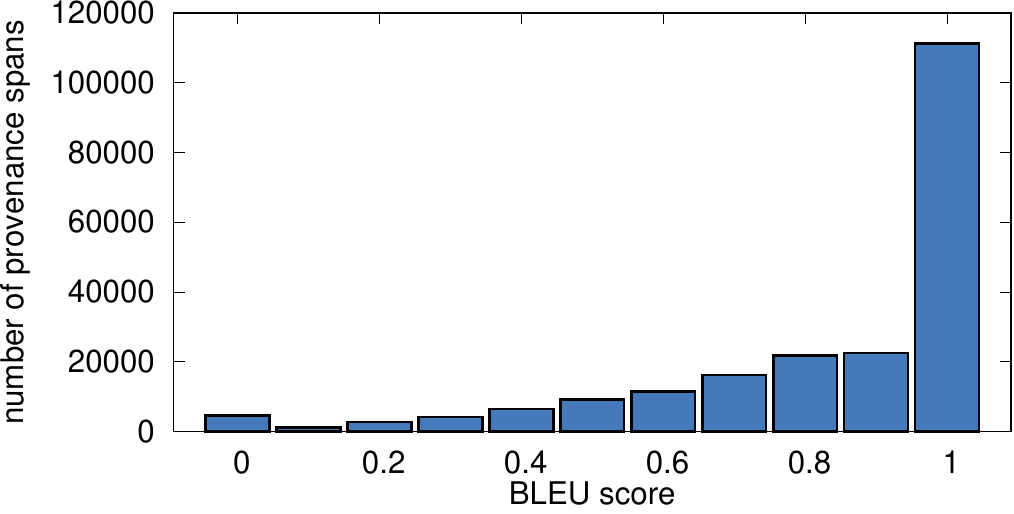}  
  \caption{HotpotQA, dev data discarded 22.76\% (1650), test data discarded 23.43\% (1704).}
  \label{fig:sub-HP}
\end{subfigure}\hspace{.05\textwidth}% Space between image B
\begin{subfigure}{.47\textwidth}
  \centering
  % include second image
  \includegraphics[width=\linewidth]{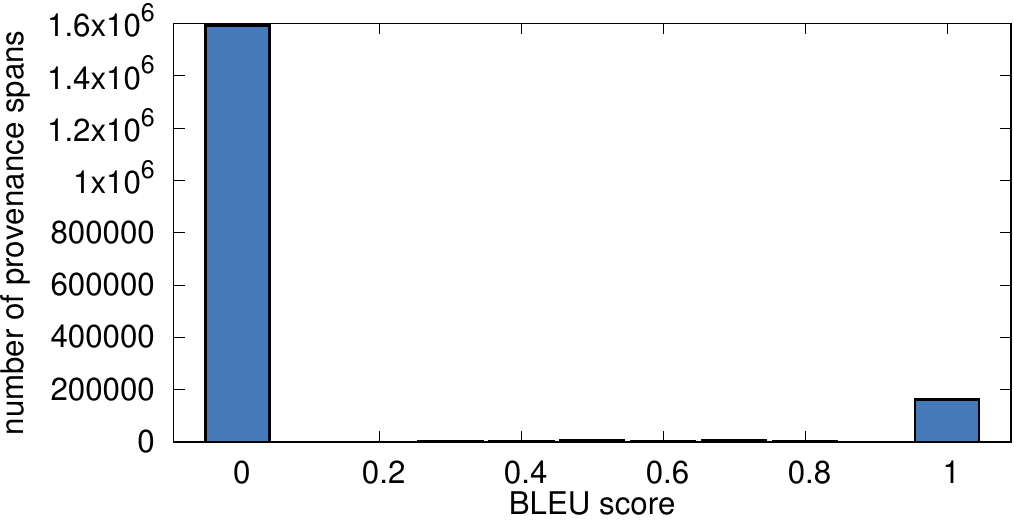}
  \caption{TriviaQA, dev data discarded 15.06\% (950), test data discarded 14.41\% (1109).}
  \label{fig:sub-TR}
\end{subfigure}

\vspace{2em}% Space between image B and C

\begin{subfigure}{.47\textwidth}
  \centering
  % include first image
  \includegraphics[width=\linewidth]{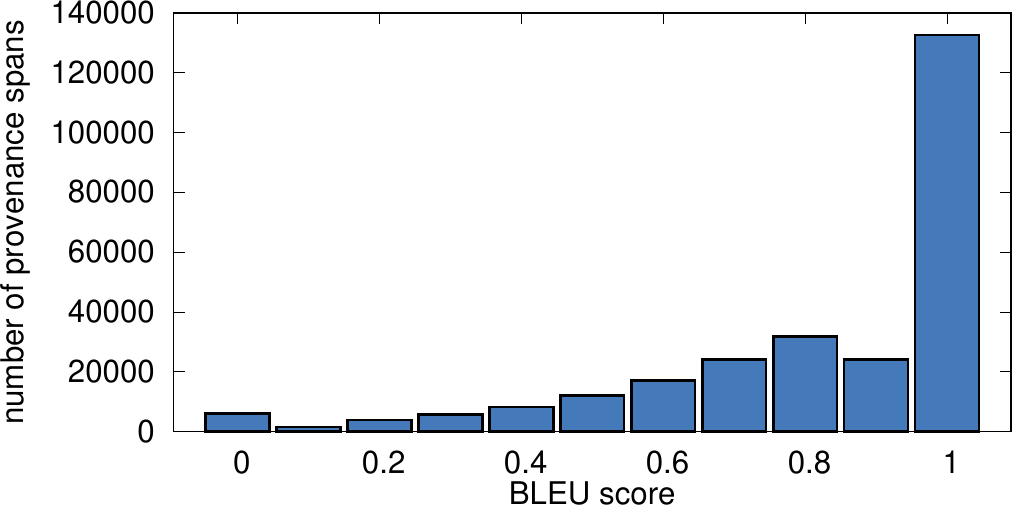}  
  \caption{Zero Shot RE, dev data discarded 15.42\% (679), test data discarded 13.38\% (767).}
  \label{fig:sub-ZE}
\end{subfigure}\hspace{.05\textwidth}% Space between image B
\begin{subfigure}{.47\textwidth}
  \centering
  % include second image
  \includegraphics[width=\linewidth]{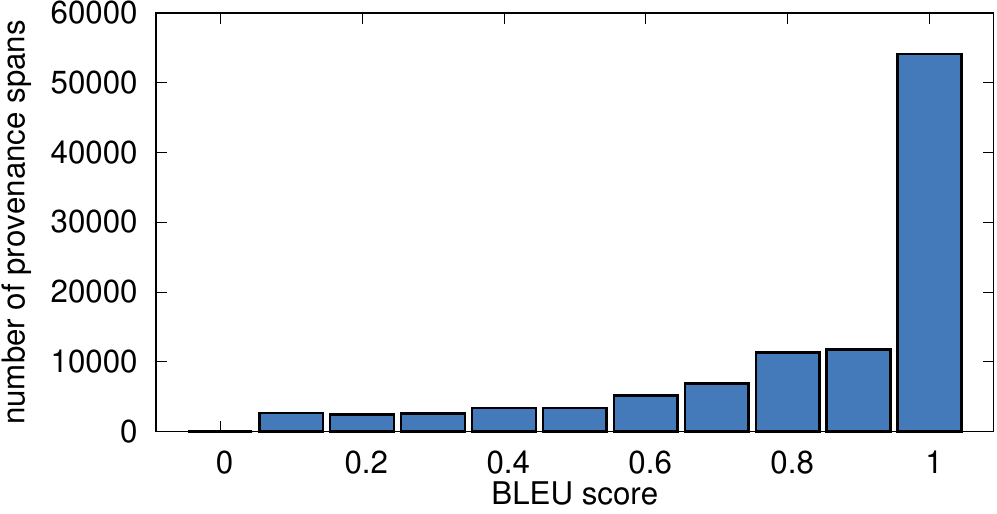}  
  \caption{Wizard of Wikipedia, dev data discarded 12.06\% (469), test data discarded 11.39\% (427).}
  \label{fig:sub-WW}
\end{subfigure}

\caption{BLEU score distribution in train data per provenance. For TriviaQA, we try to map all object aliases for the answer. FEVER has the oldest Wikipedia snapshot. We discards on average 17.9\% dev and 17.65\% test data. For TriviaQA there are a large number of 0 scores because we try to map all aliases for the answer and most of the aliases are not found in a Wikipedia page. Note that we consider a QA pair valid if we match at least one alias.}
\label{fig:bleu}
\end{figure*}

%\input{tables/datasets_joined}

%\paragraph{Discussion}

%\fabio{metric discussion}

%\fabio{the difficulty of defining a single score, around leaderboards in general?}
%There are custom solutions that can easily simplify the slot filling task. For instance, subject entities can be used for lookups by title in Wikipedia to retrieve knowledge (this heuristic will always work for zsRE), and structured human-curated resources (such as Wikidata\footnote{\url{https://www.wikidata.org}}) could be used to get all answers right. 
%Nevertheless, we are interested in testing if a general model can extract attributes about specific entities from a large body of text. 

%by manipulating the Zero Shot RE dataset. However, there are custom solutions that can easily simplify the task. For instance, subject entities can be used for lookups by title in Wikipedia to retrieve knowledge, and structured human-curated resources (such as Wikidata\footnote{\url{https://www.wikidata.org}}) could be used to get all answers right. 
%Nevertheless, we are interested in testing if a general model can extract attributes about specific entities from a large body of text. 
% Hence, we consider ZS as a sort of ``unit test'' for KILT.

%The provenance to justify each system prediction can come from anywhere, including a different system, and this is difficult to detect.
%Moreover our provenance might not be exhaustive---given the redundancy of information in Wikipedia there could be other pages with the knowledge needed to solve a KILT instance. We conduct an annotation campaign to mitigate the problem.

\end{document}